\documentclass[10pt,twocolumn,letterpaper]{article}

\usepackage[pagenumbers]{wacv} 

\usepackage{graphicx}
\usepackage{amsmath}
\usepackage{amssymb}
\usepackage{booktabs}
\usepackage{balance}

\usepackage[pagebackref,breaklinks,colorlinks]{hyperref}
\usepackage[ruled,linesnumbered]{algorithm2e}
\usepackage{tabularx}
\usepackage{csvsimple}
\usepackage{tikz}

\usepackage[capitalize]{cleveref}
\crefname{section}{Sec.}{Secs.}
\Crefname{section}{Section}{Sections}
\Crefname{table}{Table}{Tables}
\crefname{table}{Tab.}{Tabs.}

\newcommand\blfootnote[1]{
    \begingroup
    \renewcommand\thefootnote{}\footnote{#1}
    \addtocounter{footnote}{-1}
    \endgroup
}

\providecommand{\keywords}[1]{\textbf{\textit{Keywords : }} #1}

\begin{document}

\title{A High-Accuracy SSIM-based Scoring System for Coin Die Link Identification}

\author{
Patrice Labedan\\
ISAE-SUPAERO,\\
Université de Toulouse, France\\
{\tt\small patrice.labedan@isae-supaero.fr}
\and
Nicolas Drougard\\
ISAE-SUPAERO,\\
Université de Toulouse, France\\
{\tt\small nicolas.drougard@isae-supaero.fr}
\and
Alexandre Berezin\\
ISAE-SUPAERO,\\
Université de Toulouse, France\\
{\tt\small alexandrbe@hotmail.fr}
\and
Guowei Sun\\
ISAE-SUPAERO,\\
Université de Toulouse, France\\
{\tt\small joelsgw@163.com}
\and
Francis Dieulafait\\
Had\`es, Bureau d'investigations\\
arch\'eologiques, L'Union, France\\
{\tt\small francis.dieulafait@hades-archeologie.com}
}

\maketitle


\begin{abstract}
The analyses of ancient coins, and especially the identification of those struck with the same die, provides invaluable information for archaeologists and historians. Nowadays, these die links are identified manually, which makes the process laborious, if not impossible when big treasures are discovered as the number of comparisons is too large. This study introduces advances that promise to streamline and enhance archaeological coin analysis. Our contributions include: 1) First publicly accessible labeled dataset of coin pictures (329 images) for die link detection, facilitating method benchmarking; 2) Novel SSIM-based scoring method for rapid and accurate discrimination of coin pairs, outperforming current techniques used in this research field; 3) Evaluation of clustering techniques using our score, demonstrating near-perfect die link identification. 
We provide datasets \cite{AFRCBK_2024}, to foster future research and the development of even more powerful tools for archaeology, and more particularly for numismatics.
\end{abstract}


\vspace{-0.2cm}
\vspace{-0.1cm}

\section{Introduction}
\vspace{-0.3cm}
\blfootnote{Submitted to WACV 2025}

There is no doubt that Artificial Intelligence, and Machine Learning in particular, can make a major contribution to the field of archaeology \cite{barcelo2010visual,mantovan2020computerization,argyrou2022review,pavlidis2023digital}. Although there are still very few labeled archaeological data freely available online, which hinders the development and evaluation of information extraction techniques, research has been carried out into the creation of such datasets \cite{klein2023autarch,siddiqui2022metadata}.

The study of ancient coins (Ancient Numismatics) has become an attractive research field in recent years thanks to the application of Machine Learning and Computer Vision algorithms. Early works focused on the main aspect of analyzing an ancient coin, namely the identification of its issue (issuing authority, mint, etc.) from photos \cite{guo2023siamese,anwar2021deep,schlag2017ancient,arandjelovic2020images,cooper2020learning,arandjelovic2012reading,ma2020classification}. While the performance of these algorithms has improved over the years, it still falls short of the accuracy achieved by an experienced numismatist.

\subsection{Die Link Detection}

\begin{figure}
    \centering
	  \includegraphics[width=0.9\linewidth]{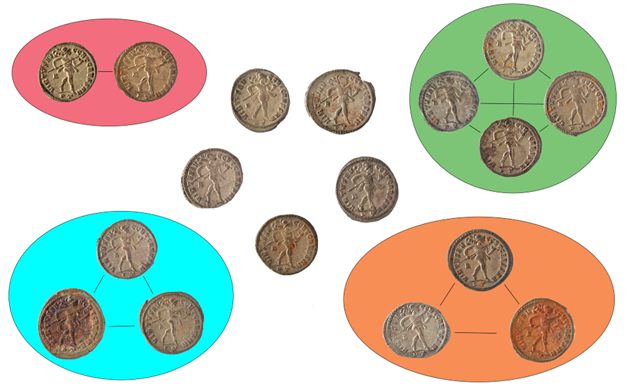}
	  \caption{Coins struck with the same die: example of ground truth on a dataset of 17 coins (DS8, as defined in Table \ref{tab:dsslc}).} 
	  \label{fig:algo_sur_ds8} 
\end{figure}

Ancient coins were produced in mints using two steel-coated iron dies — one for the obverse and one for the reverse. As these dies wear out, they are replaced, and since each die is engraved by hand, minor differences can be observed in coins struck from different dies. 

\textit{Die-linked coins}, those minted with the same obverse or reverse dies, can provide crucial information about mint organization, sequence of issues and their dating, which are of great interest to historians and archaeologists, as they help establish connections across time and space, and refining our understanding of historical events \cite{esty1990theory}.

Therefore, when studying a coin collection, such as those from hoards, numismatists often search for \textit{die links}, i.e. the identification of coins struck by the same engraved dies. The result of such a task is illustrated in Fig. \ref{fig:algo_sur_ds8}.
The traditional method of identifying these links is labor-intensive and impractical for large collections. For example, in a hoard like the one of L'Isle-Jourdain (see Fig.\ \ref{sec:dataset}), containing 1,395 coins with the most common reverse type, there are 972,315 possible pairings to examine, each requiring an average of five seconds to check \cite{doyen2019big}. This amounts to approximately 1,350 hours of work, making the analysis of large hoards almost unfeasible. However, recent studies have begun to address this challenge using Computer Vision and Machine Learning techniques, offering a more efficient approach to identifying die links in vast collections \cite{Base, Cohesion}.

Initiatives are also underway to make ancient coin data publicly available online, but there are as yet no image databases available for the automatic clustering of coin images, grouping together coins struck by the same die. Such datasets exist for classification problems \cite{aslan2020two}, as well as for clustering based on 3D scans \cite{horache2021riedones3d}, but not for die link detection from pictures.




Although it is not the case for coin classification \cite{anwar2021deep}, the small amount of publicly available labeled data also makes it more difficult to use Deep Learning techniques for automatic feature extraction. Automated analysis therefore still relies on the development of scoring techniques tailored to the datasets studied \cite{taylor2020computer, Base, Cohesion}. Thus, it is important to highlight the analysis tools that perform well in these specific tasks, and that could be used in the future to accurately label datasets.

\subsection{Contributions}

In this context, this paper presents the first labeled image dataset for evaluating clustering methods for automatic identification of coins struck with the same die, described in Section \ref{sec:dataset}. After a presentation of the related works in Section \ref{sec:rw}, a new procedure for computing distances between coins, based on the \textit{Structural Similarity Index Measure} (SSIM \cite{wang2004image}) is detailed in Section \ref{ref:ssim}, along with a state-of-the-art procedure used as a baseline in this study. Section \ref{sec:eval} highlights the superior ability of this new distance to discriminate die links compared to previous approaches in the literature. 
Finally, the results of state-of-the-art distance-based clustering algorithms are presented and analyzed, demonstrating the accuracy and efficiency of this fast distance computation technique for the study of die links.

\section{An Image Dataset for Die Link Detection}
\label{sec:dataset}
\begin{figure}[b]
    \centering
	  \includegraphics[width=0.6\linewidth]{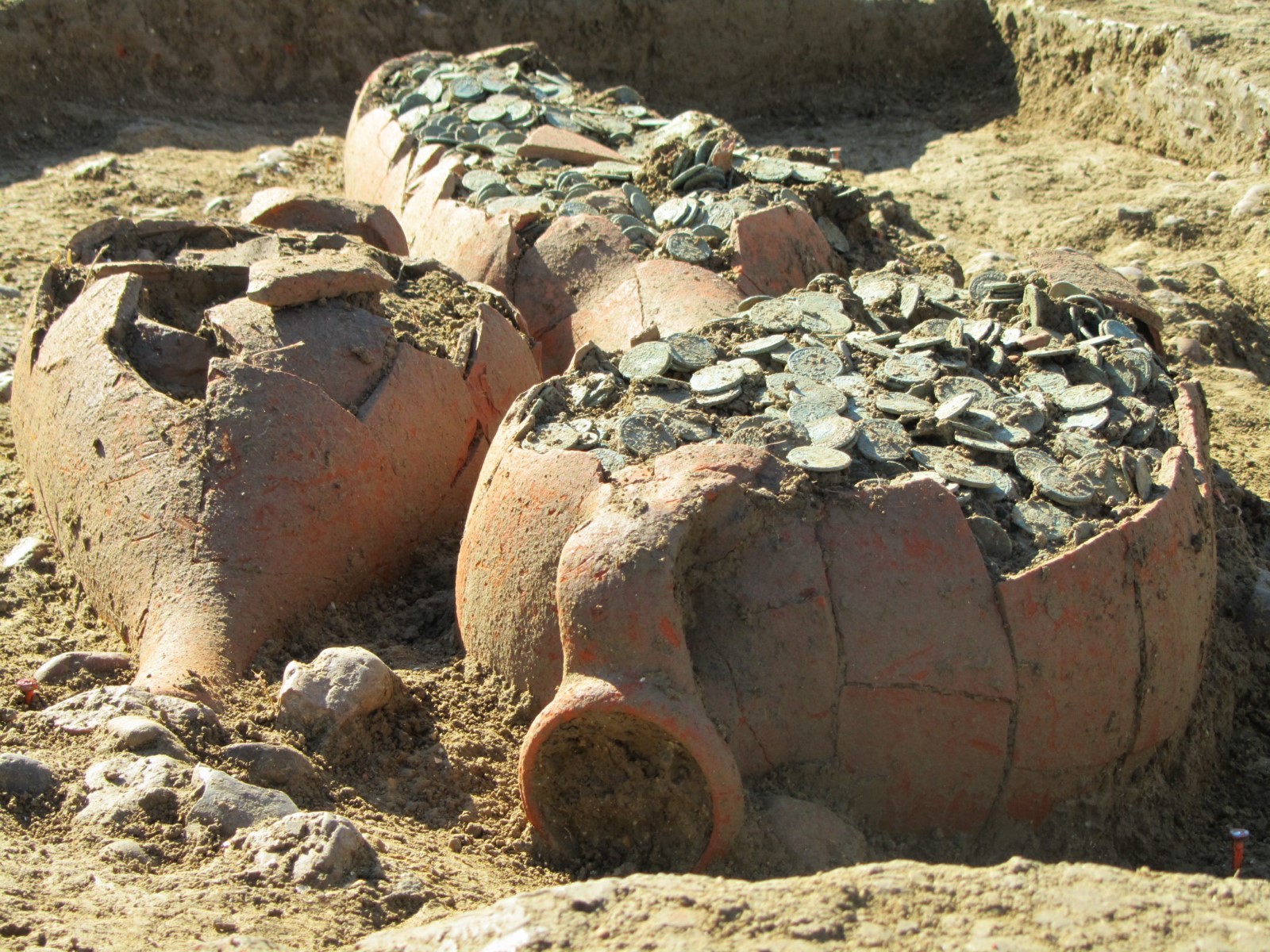}
	  \caption{The Juillac treasure during the archaeological dig.}
	  \label{fig:depot_fouille} 
\end{figure}

The Juillac treasure (Fig.\ \ref{fig:depot_fouille}) was discovered in 2011 in the municipality of L'Isle-Jourdain (Gers, France). The datasets used for our work come from the scientific study of this important treasure. It contains more than 23,200 Roman coins, mainly dated between 294 and 313 AD. The archaeologists and numismatists studying this hoard analyzed each coin, which is documented on both sides (called the obverse and reverse) with a digital photograph and several descriptive headings, six of which are used for this research
(see supplementary material). 
These six headings alone make it possible to classify all the coins by type of obverse and type of reverse. If we only keep the types composed of at least two coins, the database thus contains 658 different types of reverse (from 2 to 1,395 coins), and 379 different types of obverse (from 2 to 1,255 coins). For the study of this hoard, the numismatists created an innovative database in the field of large hoards. It allows easy access to the record of each coin and, more importantly, to the coins of each previously identified type, enabling comparisons between pairs of coins. The visual analysis of die links has thus started for certain types of obverse or reverse. At the time of our work, this is the case for coins from the \textit{Ticinum} mint, with a relatively small number of coins examined (batches containing from 2 to 93 coins). Eight sets are used as references, called here DS1, DS2, ..., DS8. The numismatists allowed us to use and publicly share these datasets \cite{AFRCBK_2024}.


\begin{table}[b]
    \begin{center}
            \footnotesize
        \begin{tabular}{cccc}
            \hline 
            Dataset  & Number   & Number of     & Number of   \\
                     & of coins & possibilities & actual links  \\
            \hline
            DS1             &  81  &  3240 &   3 \\
            DS2             &  19  &   171 &   2 \\
            DS3             &  53  &  1378 &  10 \\
            DS4             &  56  &  1540 &   3 \\
            DS5             &  49  &  1176 &   4 \\
            DS6             &  22  &   231 &   1 \\
            DS7             &  32  &   496 &   2 \\
            DS8             &  17  &   136 &  13 \\
            \hline
            Total           & 329  &  8368 &  38 \\
            \hline
        \end{tabular} 
        \caption{\label{tab:tab_datasets_used}Dataset sizes and label counts. The number of potential die links in a dataset of size $n$ is $\frac{n(n-1)}{2}$.}
        \label{tab:dsslc}
    \end{center}
\end{table}



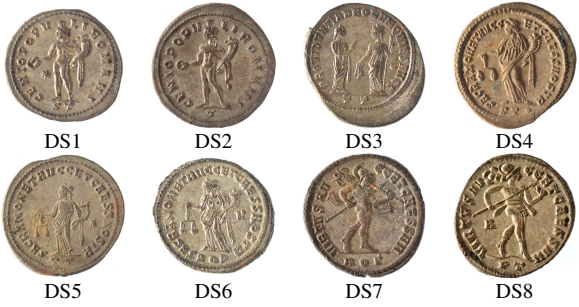
\begin{figure}
\begin{tikzpicture}
\node at (0,1) {\footnotesize DS1};
\node at (0,2) {\includegraphics[width=0.2\linewidth]{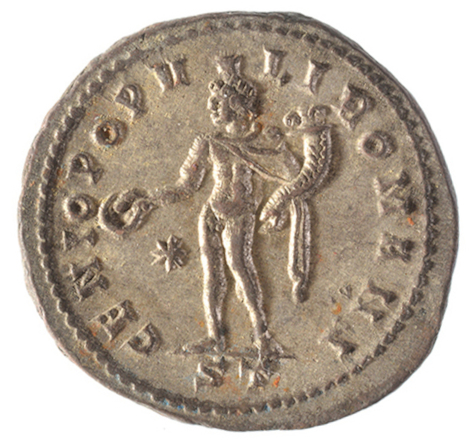}};
\node at (2,1) {\footnotesize DS2};
\node at (2,2) {\includegraphics[width=0.2\linewidth]{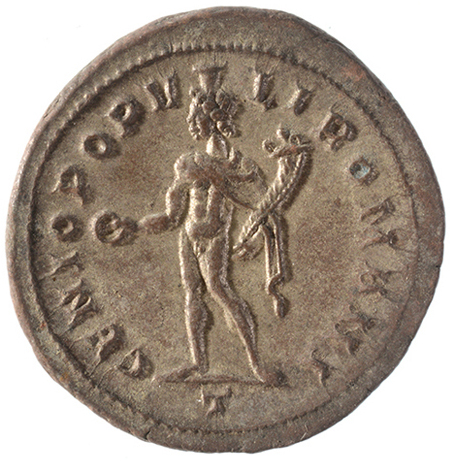}};
\node at (4,1) {\footnotesize DS3};
\node at (4,2) {\includegraphics[width=0.2\linewidth]{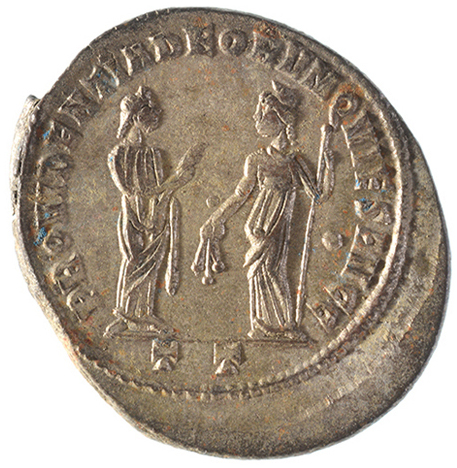}};
\node at (6,1) {\footnotesize DS4};
\node at (6,2) {\includegraphics[width=0.2\linewidth]{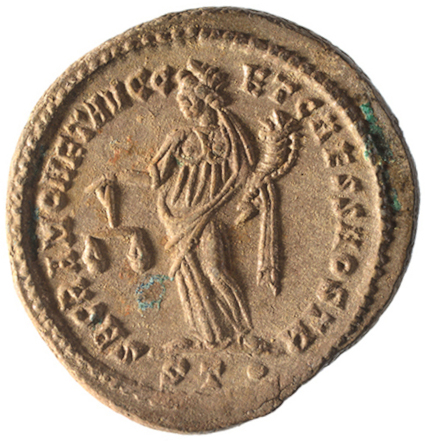}};

\node at (0,-1) {\footnotesize DS5};
\node at (0,0) {\includegraphics[width=0.2\linewidth]{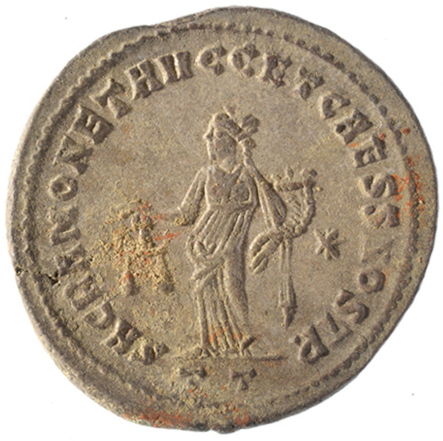}};
\node at (2,-1) {\footnotesize DS6};
\node at (2,0) {\includegraphics[width=0.2\linewidth]{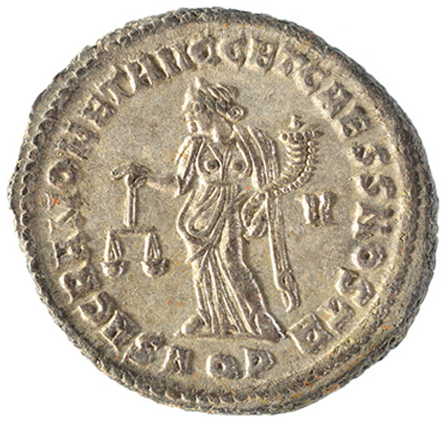}};
\node at (4,-1) {\footnotesize DS7};
\node at (4,0) {\includegraphics[width=0.2\linewidth]{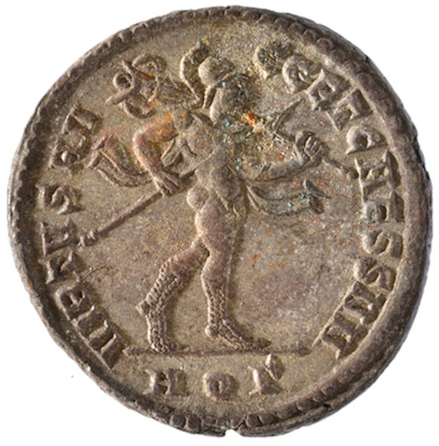}};
\node at (6,-1) {\footnotesize DS8};
\node at (6,0) {\includegraphics[width=0.2\linewidth]{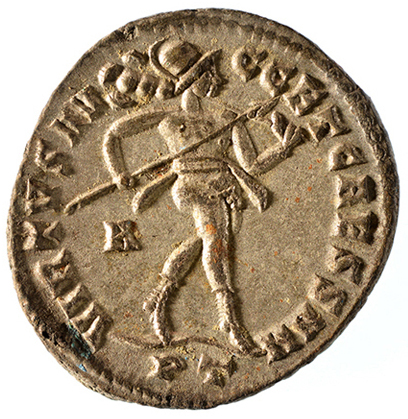}};
\end{tikzpicture}
  \caption{Coin example for each dataset}
  \label{tab:mon_tableau}
\end{figure}

The lighting conditions under which the images are taken (the 46,400 digital images in the database) have a major impact on the results of die link detection. Our algorithms are based on detecting points of interest on coins. From a numismatic point of view, good photos are taken with semi-glare lighting (to reproduce the slightest relief and legibility) and, above all, with a light source that is always to the left (i.e. aimed at the back of the emperor's neck if the coin is correctly positioned under the lens). Out of the eight datasets we kept, i.e. 401 coins, the lighting was correct for only 329. This problem stems from the fact that the photographer sometimes forgot the instructions for positioning the light source. For our purposes, we only kept the coins that were lit in the conventional way, i.e. from the left. A description of the used datasets \cite{AFRCBK_2024} is given in Table \ref{tab:tab_datasets_used}, and picture examples are given in Fig.\ \ref{tab:mon_tableau}.

The 329 images extracted from the scientific database of the treasure are each $787 \times 787$ pixels in size, featuring a resolution of 200 pixels per inch both horizontally and vertically.

\section{Related Works}
\label{sec:rw}

In the domain of coin die link detection, some recent works are very promising \cite{taylor2020computer, Base, Cohesion}.
However, the datasets used in these works are not publicly available, and the source codes for computing coin dissimilarities have not been released online. The first work \cite{taylor2020computer} uses \textit{Oriented FAST and rotated BRIEF} (ORB \cite{rublee2011orb}) to extract points of interest, also called \textit{keypoints}, from coin pictures, then brute force matching to match points between two coins, and finally averages the descriptor distances of the best matches to obtain a dissimilarity measure. The method used in \cite{Base} extracts keypoints using Gaussian processes \cite{Gaussian}, associates descriptors with VLFeat \cite{vedaldi2010vlfeat}, matches keypoints using a bounded distortion feature matching method \cite{lipman2014feature}, and finally computes a dissimilarity measure based on the Procrustes distance between these point sequences, and the number of matches. Finally, the procedure described in \cite{Cohesion} uses SIFT to obtain keypoints and descriptors, matches keypoints using the ratio test \cite{lowe2004distinctive} and bounded distortion feature matching, and finally combines the Procrustes distance, the number of matches, and the descriptors and average local gradients to construct a dissimilarity measure.

These methods computes dissimilarity measures between pictures, from local features, i.e. the keypoints and their associated descriptors. In this paper, we propose to focus on a global measure of the similarity between images, based on SSIM, in order to benefit from all the information contained in the images when comparing them.

\section{SSIM-based distance}
\label{ref:ssim}
The dissimilarity measures used for this problem in the literature, e.g. based on Procrustes distance, indicate how numerous and similar the paired keypoints are, and how overlapping they can be. In other words, once the points of interest have been extracted, they are sufficient to compute the distance, and no additional information from the images is used. The main idea behind the distance presented in this paper is to continue to take advantage of the information in the images when computing the dissimilarity values. Once the images have been superimposed, the structural similarity index between the two images is computed, taking into account all the image details.

The structural similarity (SSIM) index \cite{wang2004image, nilsson2020understanding} can be defined as a function of two images $A,B \in R_{+}^{n \times m}$ returning an image with the same size: $\forall (i,j) \in \{1,\ldots,n\} \times \{1, \ldots, m\}$,
$S_{ij}(A,B) = \big(l^{AB}_{ij}\big)^{\alpha} \big(c^{AB}_{ij}\big)^{\beta} \big(s^{AB}_{ij}\big)^{\gamma}$,
\textit{i.e.} the product of three components, namely \textit{luminance}, \textit{contrast} and \textit{structure}.

The \textit{luminance} component $l^{AB}_{ij} = \frac{2 \mu^A_{ij}\mu^B_{ij} + c_1}{(\mu^A_{ij})^2 + (\mu^B_{ij})^2 + c_1}$ depends on the \textit{local} means $\mu^A_{ij}$ and $\mu^B_{ij}$ \textit{i.e.} the means computed in a patch around pixel $(i,j)$ with gaussian weights \cite{nilsson2020understanding}. The contrast component $c^{AB}_{ij} = \frac{2 \sigma^A_{ij}\sigma^B_{ij} + c_2}{(\sigma^A_{ij})^2 + (\sigma^B_{ij})^2 + c_2}$ depends on the local standard deviations $\sigma^A_{ij}$ and $\sigma^B_{ij}$. Finally, the \textit{structure} component $s^{AB}_{ij} = \frac{\sigma_{ij}^{AB} + c_3}{\sigma_{ij}^{A}\sigma_{ij}^{B} + c_3}$ depends on the local covariance $\sigma_{ij}^{AB}$. In the sake of simplicity, the following constants have been finally chosen: $\alpha = \beta = \gamma = 1$ and $c_3 = \frac{c_2}{2}$ \cite{wang2004image}.

The SSIM index was developped for image quality assessment based on a reference image. In practice the \textit{Mean SSIM} (MSSIM) index is used to evaluate the similarity between images $A$ and $B$:
\begin{equation}
S(A,B) = \frac{1}{nm}\sum_{i\leq n, j\leq m} S_{ij}(A,B).
\label{S}
\end{equation}
This index satisfies by construction the following properties: it is symmetric, \textit{i.e.} $S(A,B) = S(B, A)$, bounded by $1$, \textit{i.e.} $\lvert S(A, B) \rvert  \leq 1$, and equal to $1$ only if $A=B$.  The MSSIM index is equal to $1$ when the input images are the same, and $-1$ when they are perfectly anti-correlated.

In order to transform this index into a dissimilarity measure, a first idea could be to use an decreasing function of SSIM, as $1-S(A,B) \in [0,2]$ for instance. However, this dissimilarity is not a metric (in the mathematical sense), \textit{i.e.} a distance function, since it does not respect the triangular inequality. This limitation can degrade the performance of distance-based clustering algorithms \cite{baraty2011impact,saha2020non,saha2024study} to be used for die link analysis. The work developed in \cite{brunet2011mathematical} defines a function very similar to the previous one, which has the advantage of being a metric, or distance function: $\forall (i,j) \in \{ 1, \ldots, n \} \times \{ 1, \ldots, m \}$,
\begin{equation}
M_{ij}(A,B) = \sqrt{2 - l^{AB}_{ij} - s^{AB}_{ij} c^{AB}_{ij}},
\label{lM}
\end{equation}
that can be seen as a low-order estimation of $\sqrt{2-S_{ij}(A,B)}$. In the same way as Equation \ref{S}, the distance function used in our work is thus defined as follows:
\begin{equation}
M(A,B) = \frac{1}{nm} \sum_{i\leq n, j\leq m} M_{ij}(A,B),
\label{M}
\end{equation}
where $M_{ij}(A,B)$ is the local SSIM distance function defined in Equation \ref{lM}.

\begin{figure}[b]
    \centering
	  \includegraphics[width=1\linewidth]{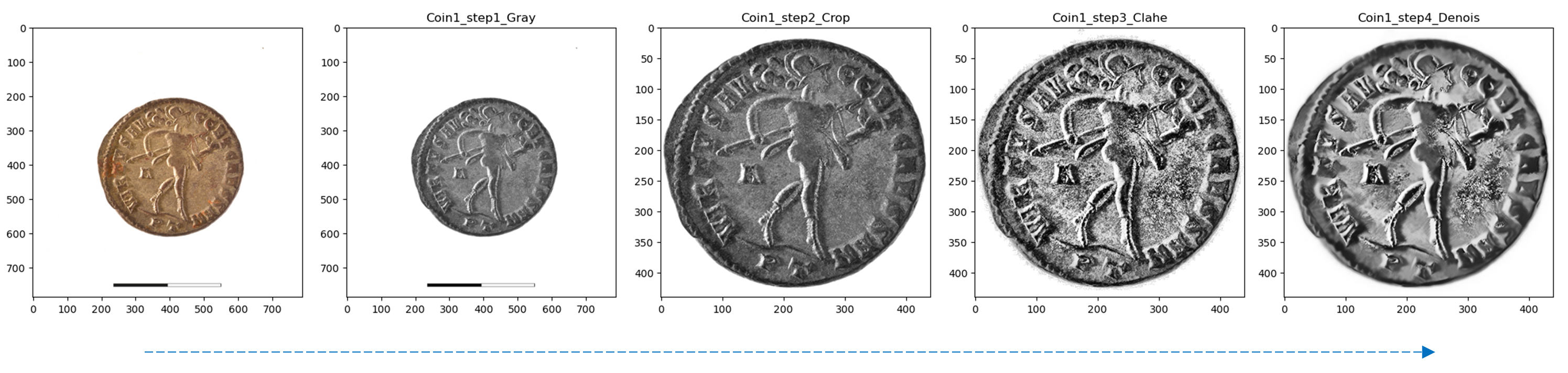}
	  \caption{Pre-processing steps from the original coin to the input of the SSIM-based method (from left to right: raw database image; grayscale; cropping; CLAHE; denoising).}
	  \label{fig:etapes_processing} 
\end{figure}

\SetKwComment{Comment}{/* }{ */}
\begin{algorithm}[t]
\caption{SSIM-based Distance Computation}
\label{algo:ssim}
\KwData{$(A,B) \in \mathbb{R}_{+}^{n\times m \times 2}$ \Comment*[r]{Images}}
\KwResult{$d_S(A,B)$ \Comment*[r]{SSIM metric}}
\For{$C \in \{ A, B \}$}{
$C \gets \texttt{preproc}_1(C)$ \Comment*[r]{B\&W, Crop, CLAHE, Fast Non-Local Means}
$(\delta^C, \kappa^C) \gets \texttt{SIFT}(C)$  \Comment*[r]{Computation of descriptors \& keypoints}
}
$(\kappa^A,\kappa^B) \gets \texttt{matcher}_1(\delta^A,\kappa^A,\delta^B, \kappa^B)$ \Comment*[r]{Brute force with ratio test}
\If{$K > 4$}{
$(s,\theta, t_x, t_y) \gets \texttt{AffTransfEstim}(\kappa^A,\kappa^B)$ \Comment*[r]{Estimate 2D transf.}
}
\If{$\lvert s -1 \rvert > 0.25$\Comment*[r]{Wrong estimation}}{
$(s,\theta, t_x, t_y) \gets (1,0,0,0)$\;
}
$A \gets \texttt{affineTransf}\big(A,(s,\theta,t_1, t_2)\big)$ \Comment*[r]{Apply 2D transformation}
$d_S(A,B) \gets M(A,B)$ \Comment*[r]{Using Eq.\ \ref{M}}
\end{algorithm}

The complete procedure for computing the SSIM-based distances from raw images is described in Algorithm \ref{algo:ssim}. First, some preprocessing is applied to each image (line 2 and Fig.\ \ref{fig:etapes_processing}): they are first grayscaled, and cropped circularly with respect to the mass center of the coin pixels. The images then go through \textit{Contrast Limited Adaptive Histogram Equalization} (CLAHE, \cite{pizer1987adaptive, reza2004realization}), and finally through \textit{Non-Local Means Denoising} \cite{buades2005non}.

After this preprocessing step, SIFT \cite{lowe1999object} descriptors of each image are computed, and a brute-force matcher with a ratio test is performed (line 5). If they are more than $4$, the matched pixels are used to estimate the 2D transformation to use to superpose them (lines 6 and 7). If the scaling $s$ of the transformation estimation is too far from $1$, the estimation is considered as wrong, and the transformation is set to identity (lines 9--11). Finally, the transformation is applied to the first image, and the distance defined in Equation \ref{M} is computed and returned (lines
13--14).

\begin{figure*}
    \centering	  \includegraphics[width=0.8\linewidth]{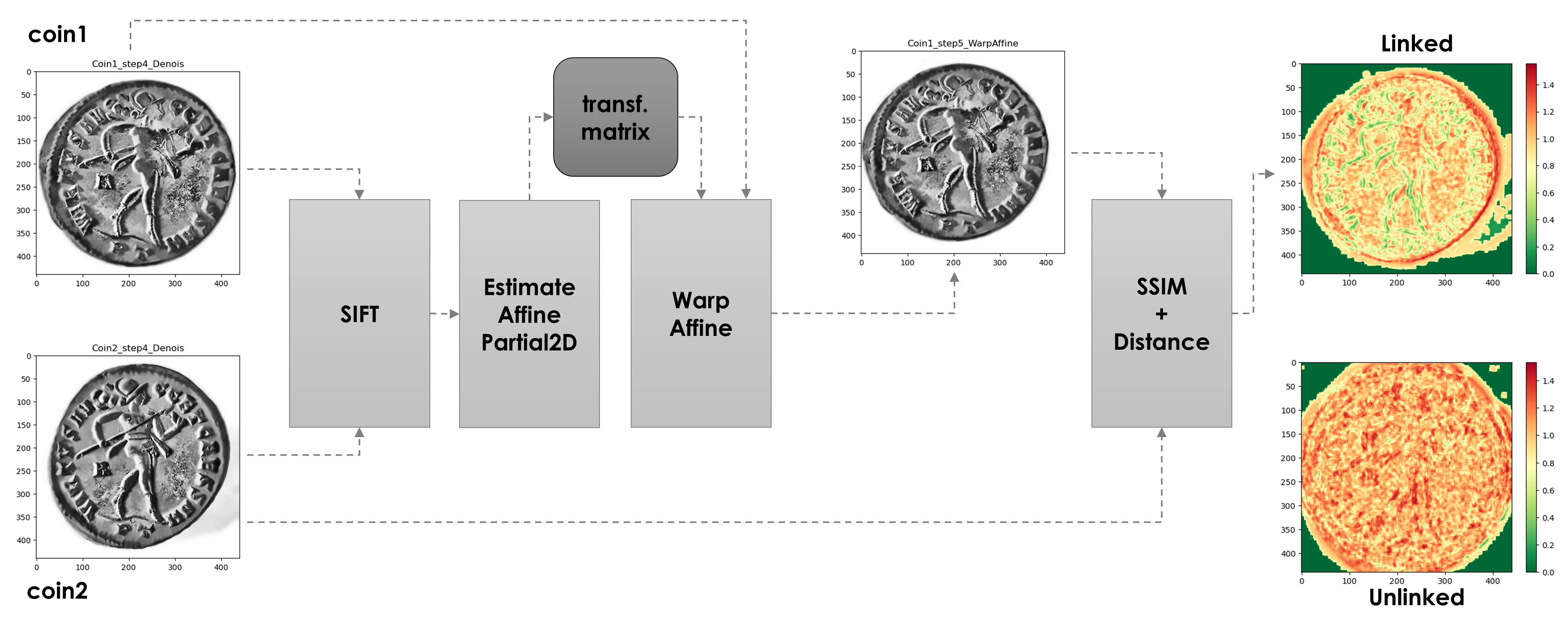}
	  \caption{Steps of the computation of the SSIM-based distance, and comparison of two results (one linked, one unlinked). The greener the color (low distance), the more likely it is to be a die link. The related algorithm is detailed in Algorithm \ref{algo:ssim}.}	  \label{fig:ssim_steps_and_result_exemple} 
\end{figure*}

The general idea of this new procedure, is to use keypoints only to allow image overlay, and then use the global SSIM score of both images, considering the entire coin surfaces, to better discriminate similarities between images (see Fig. \ref{fig:ssim_steps_and_result_exemple}).

In the next section, this new distance (using default parameters of the respective library functions to ensure a fair evaluation) is evaluated on the dataset presented in Section \ref{sec:dataset}, using as a baseline the method obtaining the best results on our datasets by reproducing the work in \cite{taylor2020computer,Base,Cohesion}. Algorithm \ref{algo:procrustes} gives the implementation details of the baseline distance computation. This distance is therefore referred to as Procrustes-based in the remainder of this article.

\SetKwComment{Comment}{/* }{ */}
\begin{algorithm}
\caption{Procrustes-based Distance Computation (baseline)}
\label{algo:procrustes}
\KwData{$(A,B) \in \mathbb{R}_{+}^{n\times m \times 2}$ \Comment*[r]{Images}}
\KwResult{$d_P(A,B)$ \hspace{-0.2cm} \Comment*[r]{Procrustes dist}}
\For{$C \in \{ A, B \}$}{
$C \gets \texttt{preproc}_2(C)$ \Comment*[r]{B\&W, Crop, TVD, CLAHE, TVD, Sobel, Crop}
$\kappa^C \gets \texttt{keypoints}(C)$ \Comment*[r]{GP \cite{Gaussian}}
$\delta^C \gets \texttt{ORB}(\kappa^C)$ \Comment*[r]{Descriptors}
}
$(\kappa^A,\kappa^B) \gets \texttt{matcher}_2(\delta^A,\kappa^A,\delta^B, \kappa^B)$ \Comment*[r]{Brute force with cross check}
$(H, n_{in}) \gets \texttt{homogrEstim}(\kappa^A, \kappa^B)$ \Comment*[r]{RANSAC-based method}
$\kappa^A \gets \texttt{homography}(\kappa^A,H)$ \Comment*[r]{Apply the estimated 3D transformation}
$d_P(A,B) \gets \log\left(P(\kappa^A,\kappa^B)\right) + \frac{1}{n_{in}}$ \Comment*[r]{$P$ is defined in Equation \ref{P}}
\end{algorithm}

The computation of the Procrustes-based distance starts with a sequence of preprocessing steps on both images, including a grayscale processing, a centered circular crop, a Total Variation Denoising (TVD, \cite{chambolle2004algorithm}), a Contrast Limited Adaptive Histogram Equalization (CLAHE, \cite{pizer1987adaptive, reza2004realization}), another TVD, a Sobel filter \cite{sobel2014history}, and a final centered circular crop (see line 2 in Algorithm \ref{algo:procrustes}). Next, keypoints (or landmarks) $\kappa^C \in \mathbb{R}^{N \times 2}$ for $C \in \{ A, B \}$, are extracted using a method based on Gaussian Processes \cite{Gaussian} (line 3), and descriptors are associated to these points using \textit{Oriented FAST and Rotated BRIEF} (ORB \cite{rublee2011orb}, line 4). Both descriptors sets are matched, cross checking to only return consistent pairs in $\kappa_A$ and $\kappa_B$ having thus a smaller size $N$ (line 6). Then, the parameters $H \in \mathbb{R}^{3 \times 3}$ of an homography ($8$ degrees of freedom) mapping $\kappa_A$ to $\kappa_B$ is estimated using \textit{Random Sample Consensus} (RANSAC, \cite{fischler1981random}), and then applied to $\kappa_A$ (lines 7 and 8). The number of inliers $n_{in} \leq N$ is also saved for the final formula. 
Finally, the distance defined as the sum of the logarithm of the \textit{Procrustes Distance} and the inverse of the number of homography inliers $n_{in}$ (line 9). In practice, this value is divided by the maximum distance value, so that the resulting distance is between zero and one.

The Procrustes Distance \cite{schonemann1966generalized}, is defined as
\begin{equation}
P(\kappa^A,\kappa^B) = \displaystyle \min_{\substack{T \in \mathbb{R}^{2 \times 2}\\
T^{\intercal}T = TT^{\intercal} = s^2I,\\ s \in \mathbb{R}}} \left\lVert \widetilde{\kappa^A} T - \widetilde{\kappa^B} \right\rVert^2, 
\label{P}
\end{equation}
where $\widetilde{\kappa^C} = \frac{\kappa^C - \overline{\kappa^C}}{\lVert \kappa^C - \overline{\kappa^C} \rVert} \in \mathbb{R}^{N \times 2}$, for $C \in \{A,B\}$, are the standardized  keypoint matrices, with $\overline{\kappa^C} \in \mathbb{R}^{N \times 2}$ such that $\forall i \in \{1, \ldots, N \}$, $\overline{\kappa^C_{i1}} = \frac{1}{N}\sum_{i'=1}^N \kappa^C_{i'1}$, $\overline{\kappa^C_{i2}} = \frac{1}{N}\sum_{i'=1}^N \kappa^C_{i'2}$, and $\lVert \kappa^C \rVert^2 = \sum_{ij} (\kappa^C_{ij})^2 $ (Frobenius norm). 
This formula 
minimizes the pointwise squared error between the transformed standardized keypoints of image $A$,  \textit{i.e.} $\widetilde{\kappa^A}T$, and the standardized keypoints of image $B$, \textit{i.e.} $\widetilde{\kappa^B}$. The set of transformations considered for this minimization, are those whose matrix representation is the product of an orthogonal matrix $Q \in \mathbb{R}^{2 \times 2}$, and a scalar $s \in \mathbb{R}$: $T=sQ$. In simpler terms, considered transformations are rotations, reflections, uniformly scaling, and combinations of these transformations. In a nutshell, this distance computes the minimal squared error of the points described by the normalized key point matrices, that can be obtained by using the mentionned transformations, as well as translations (taken into account when centering matrices $\kappa^A$ and $\kappa^B$). It can then be interpreted as a measure of ``global matching'' of these key point pairs.

Now that the baseline distance inpired by the state of the art methods (Procrustes-based, Algorithm \ref{algo:procrustes}) and the new distance introduced in this article (SSIM-based, Algorithm \ref{algo:ssim}), have been defined, it is now time to evaluate their die link identification capabilities on the provided dataset (Section \ref{sec:dataset}). Two additional methods are used in this evaluation: a distance computation based on a variant of SSIM, namely \textit{Feature SIMilarity} (FSIM, \cite{zhang2011fsim}) and another based on pre-trained deep networks, namely VGG \cite{simonyan2014very}.

\section{Distance Quality Evaluation}
\label{sec:eval}
This section is dedicated to the evaluation of the distances defined in the previous section, on the datasets described in Section \ref{sec:dataset}. 
Firstly, ROC curves and precision-recall curves as well as the areas under the ROC and PR curves (ROC AUC and PR AUC) are also computed, to assess the ability of the presented distances to detect die links. Secondly, the distributions of distance values are estimated using two histograms: one histogram for distance values representing a true die link between two parts, and another for the other distance values, which represent pairs of coins with no die link. Finally, the performances of distance-based clustering algorithms are evaluated with clustering and binary classification metrics.

In order to increase confidence in the discriminatory power of the SSIM-based distance, the default parameters have been used for all the functions from Scikit-image \cite{scikit-image} and OpenCV \cite{opencvlibrary} for the preprocessing as well as the following stages. On the opposite, Algorithm \ref{algo:procrustes} is the best possible pipeline inspired by \cite{Base,Cohesion, taylor2020computer}.
To challenge these methods using Deep Learning, image features were extracted using the pre-trained network VGG11 \cite{simonyan2014very} implemented in the Pytorch library \cite{paszke2019pytorch}, and the resulting distance between images was defined as the cosine distance between the feature vectors: $d_{\cos}(x,y) = 1-\frac{\langle x, y \rangle}{\lVert x \rVert \lVert y \rVert}$. Moreover, the  feature-similarity (FSIM, \cite{zhang2011fsim}) index, a measure comparing the low-level feature sets between the images is also used to compute a new distance for this benchmark.

\begin{figure}
    \centering 
\begin{subfigure}{0.22\textwidth}  \includegraphics[width=\linewidth]{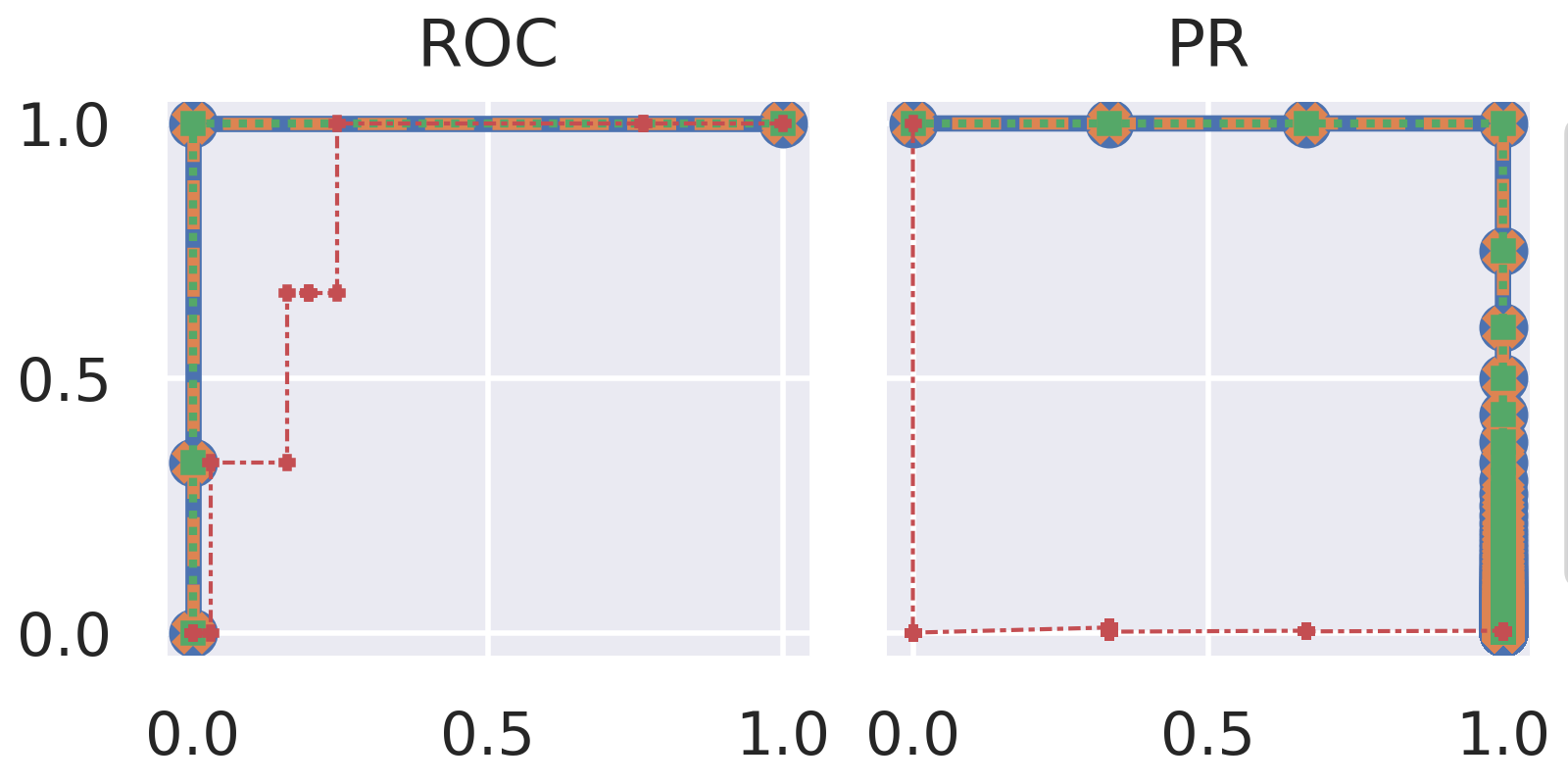}
  \caption{Dataset 1}
  \label{fig:1}
\end{subfigure}\hfil 
\begin{subfigure}{0.22\textwidth}
\includegraphics[width=\linewidth]{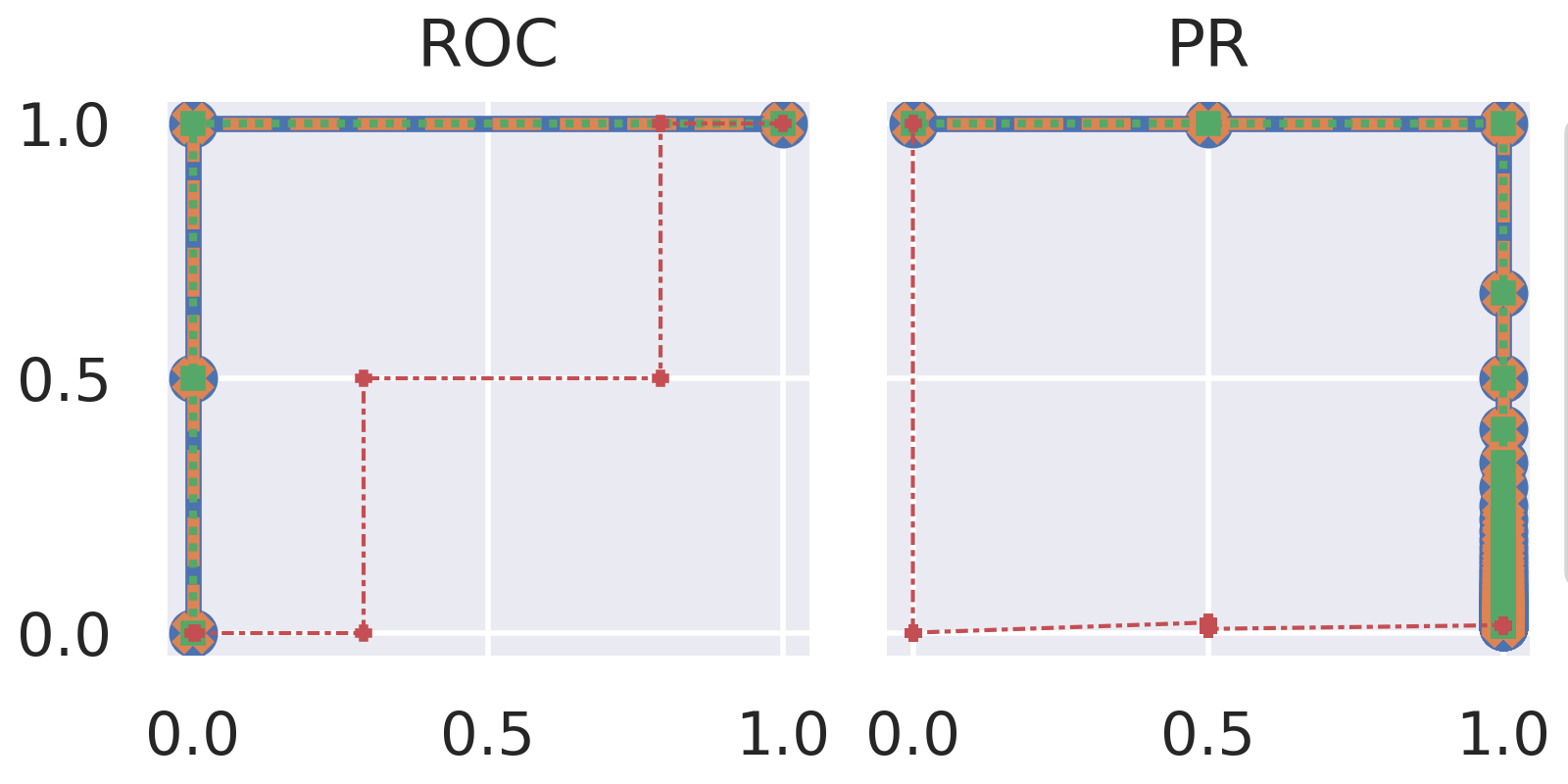}
  \caption{Dataset 2}
  \label{fig:2}
\end{subfigure}\hfil 
\begin{subfigure}{0.22\textwidth}  \includegraphics[width=\linewidth]{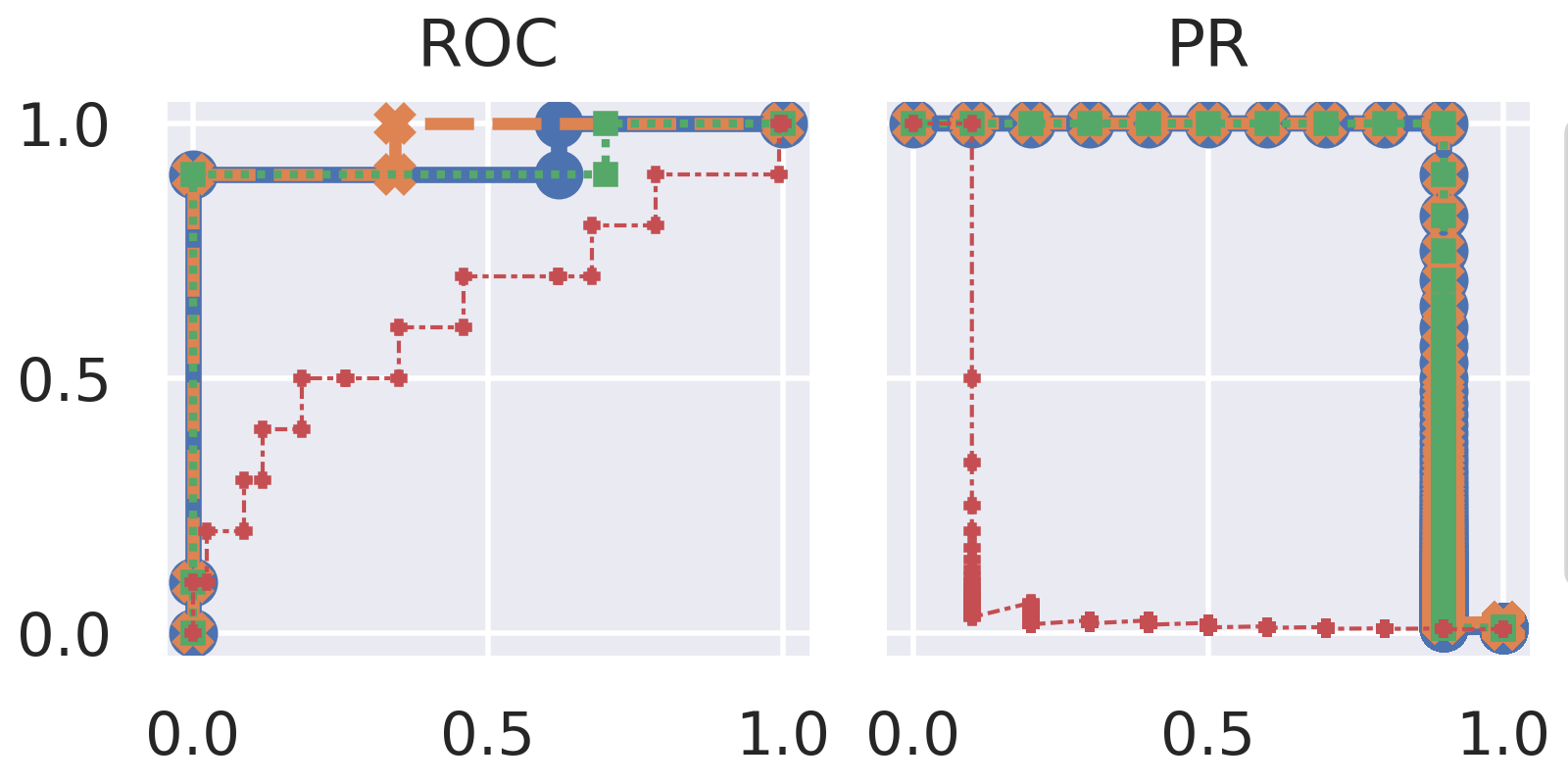}
  \caption{Dataset 3}
  \label{fig:3}
\end{subfigure}\hfil 
\begin{subfigure}{0.22\textwidth}
\includegraphics[width=\linewidth]{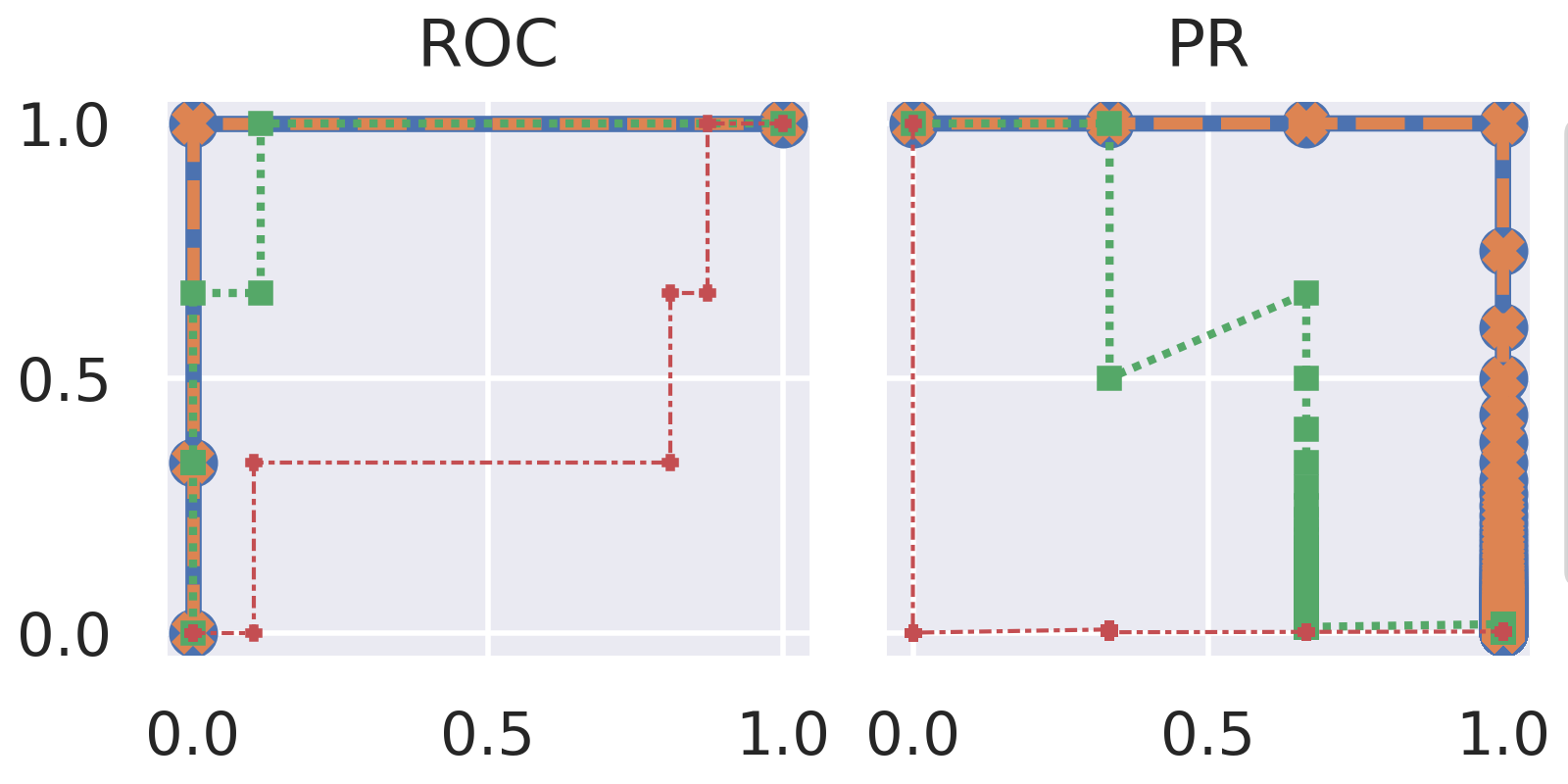}
  \caption{Dataset 4}
  \label{fig:4}
\end{subfigure}\hfil 
\medskip
\begin{subfigure}{0.22\textwidth}  \includegraphics[width=\linewidth]{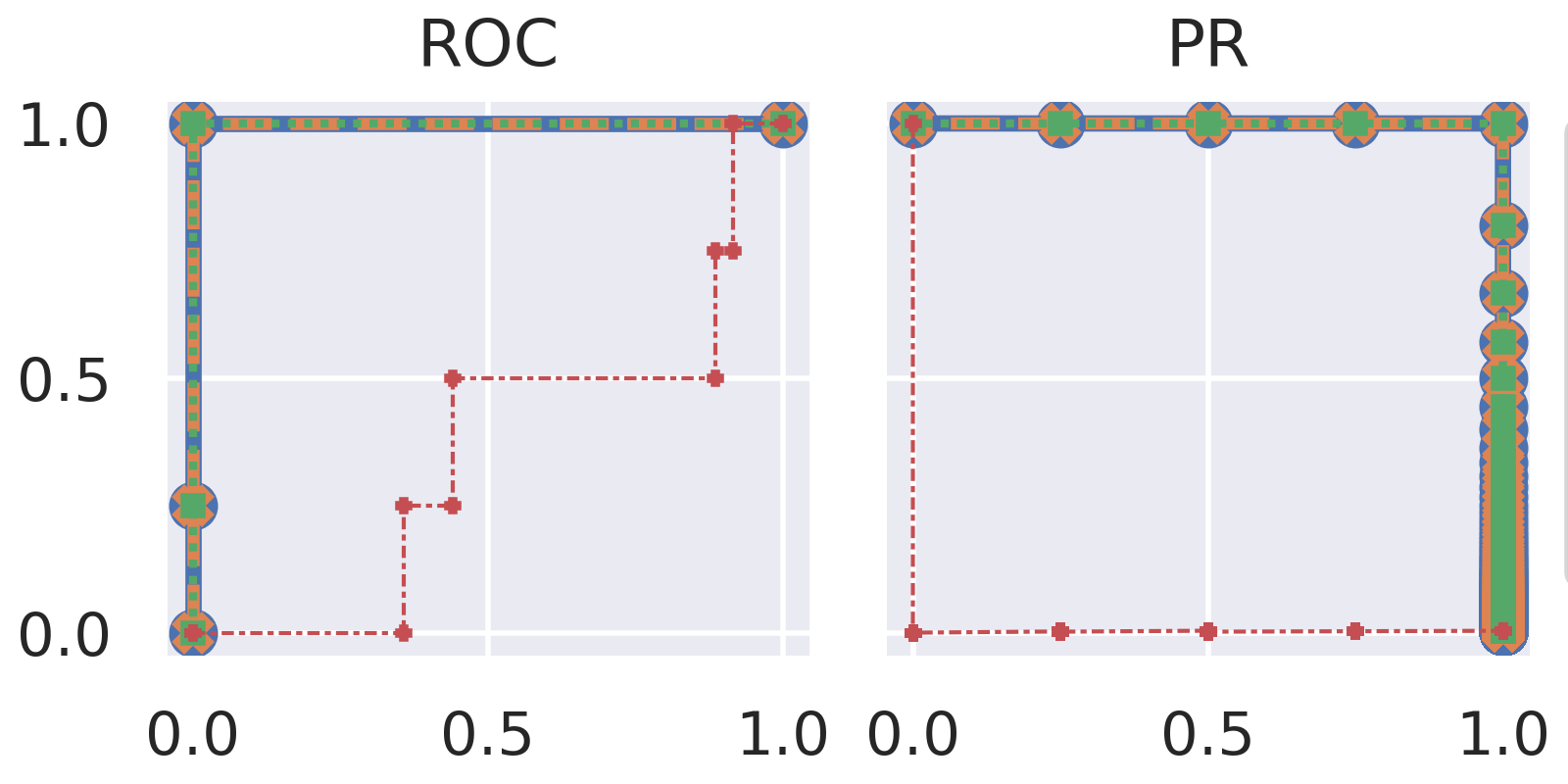}
  \caption{Dataset 5}
  \label{fig:5}
\end{subfigure}\hfil 
\begin{subfigure}{0.22\textwidth}
\includegraphics[width=\linewidth]{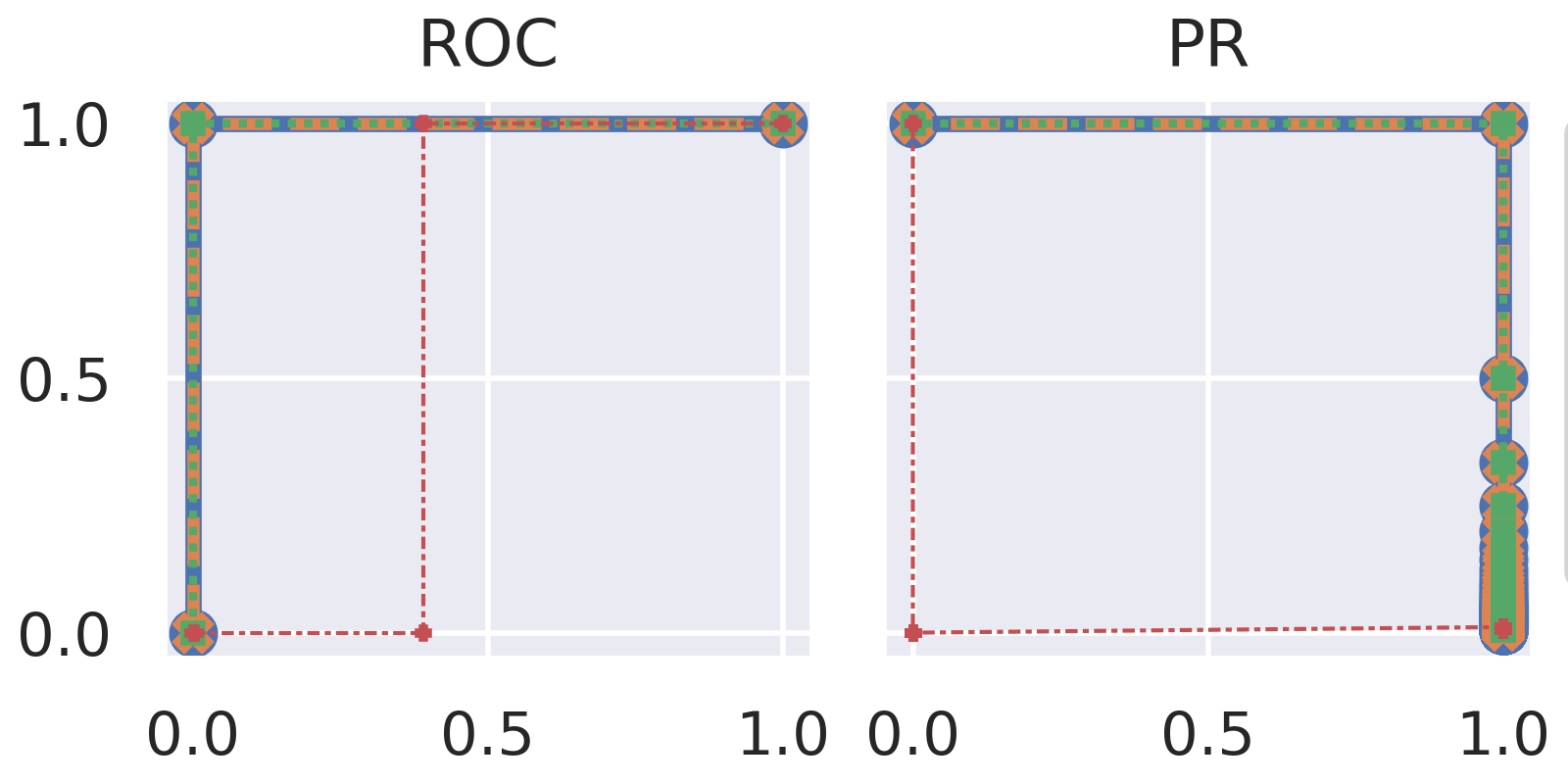}
  \caption{Dataset 6}
  \label{fig:6}
\end{subfigure}\hfil 
\begin{subfigure}{0.2\textwidth}  \includegraphics[width=\linewidth]{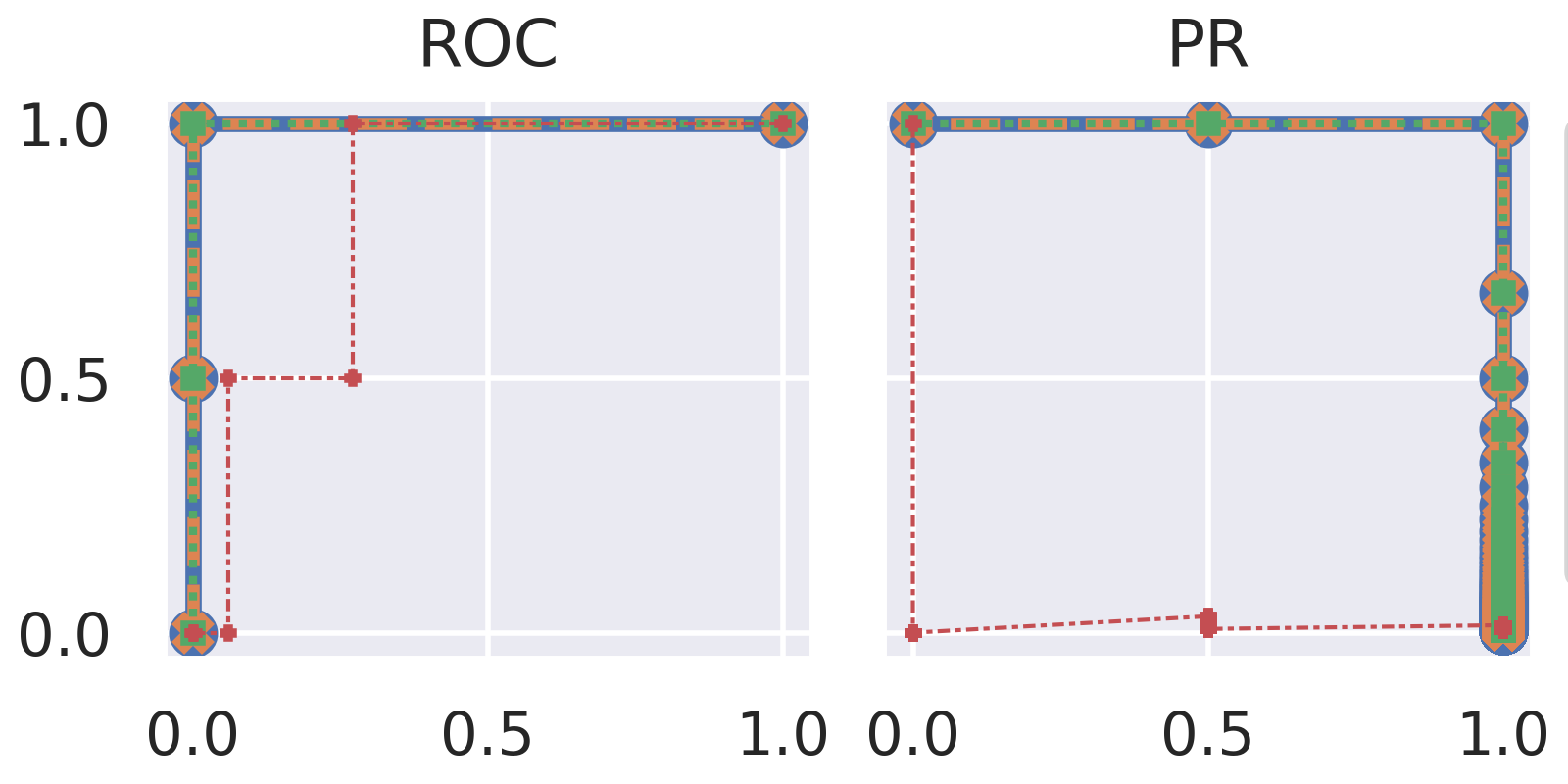}
  \caption{Dataset 7}
  \label{fig:7}
\end{subfigure}\hfil 
\begin{subfigure}{0.22\textwidth}
\includegraphics[width=\linewidth]{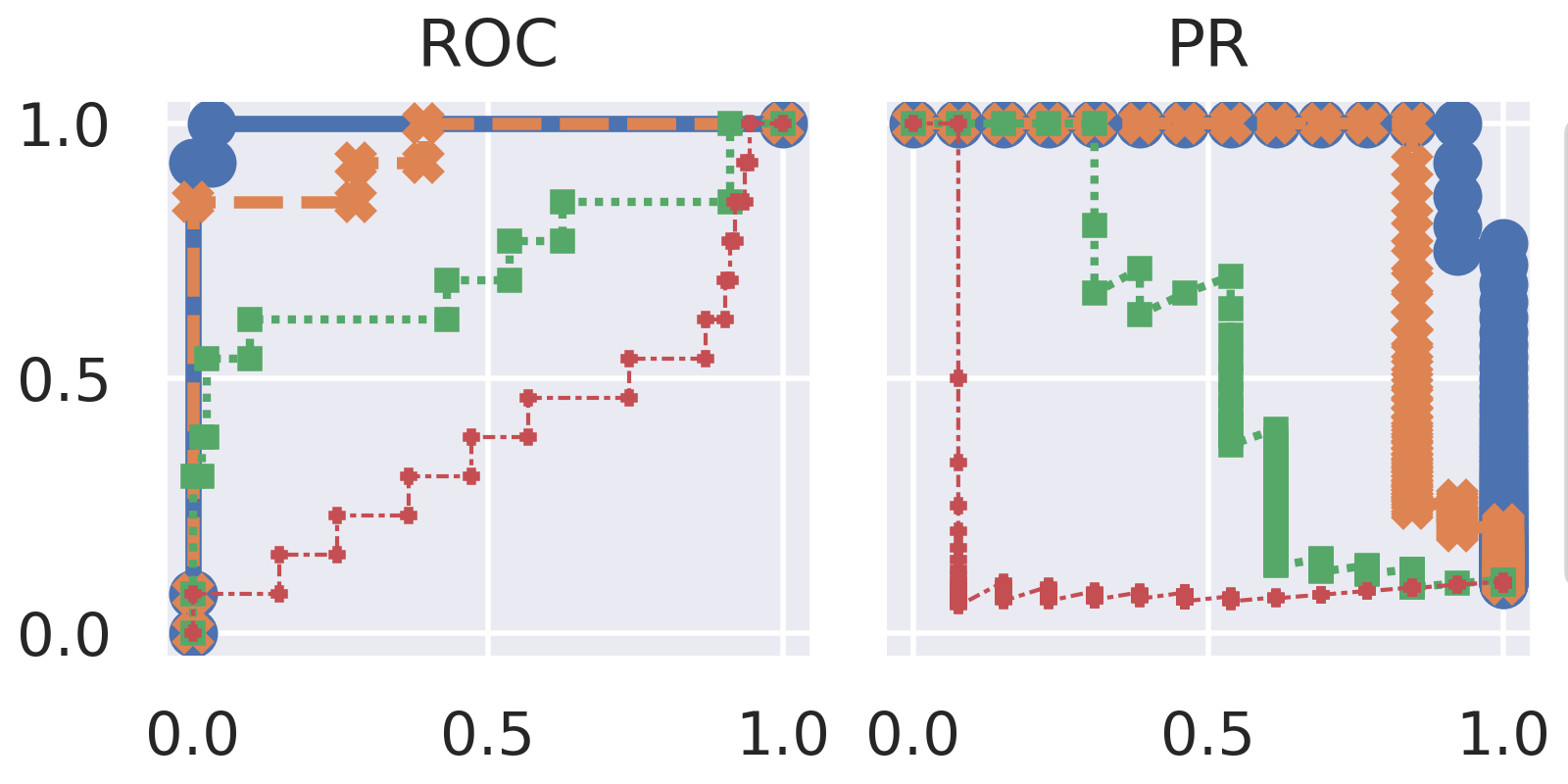}
  \caption{Dataset 8}
  \label{fig:8}
\end{subfigure}\hfil 
\includegraphics[width=0.35\linewidth]{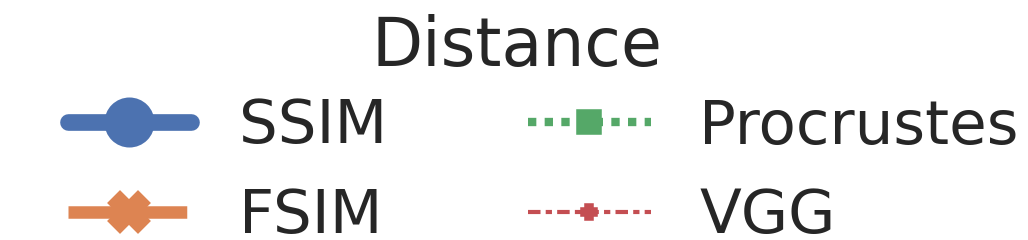}
\caption{ROC \& PR curves: SSIM-based distance in blue, FSIM-based distance in orange, Procruste-base distance in blue and VGG-based distance in red.}
\label{fig:rocr}
\end{figure}

The ROC and Precision-Recall curves associated with these distances are shown in Fig.\ \ref{fig:rocr}.
Five datasets (1, 2, 5, 6 and 7) are perfectly handled by all distances except the VGG-based distance, while the SSIM- and FSIM-based distances also obtain perfect curves for the fourth dataset. Datasets 3 and 8 seem more difficult to process, but the ROC and PR curves confirm the quality of the SSIM- and FSIM-based distances, followed closely by the Procrustes-based distance, and finally the poorer performance of the VGG-based distance.

The areas under the ROC and precision-recall curves (ROC and PR AUC), presented in tables \ref{tab:resaucroc} and \ref{tab:resaucpr}, summarise these results. As can be seen from the curves, the FSIM-based distance performs best on the third dataset, and the SSIM-based distance performs best on the eighth dataset.

\begin{table}[b]
    \begin{center}
    \footnotesize
        \begin{tabular}{ccccc} 
            \hline 
            DS  & SSIM   & FSIM & Procrustes & VGG \\
            \hline
            1 & 1.0 & 1.0 & 1.0 & 0.855 \\
            2 & 1.0 & 1.0 & 1.0 & 0.459 \\
            3 & 0.938 & \textbf{0.966} & 0.93 & 0.632 \\
            4 & 1.0 & 1.0 & 0.961 & 0.405 \\
            5 & 1.0 & 1.0 & 1.0 & 0.35 \\
            6 & 1.0 & 1.0 & 1.0 & 0.609 \\
            7 & 1.0 & 1.0 & 1.0 & 0.834 \\
            8 & \textbf{0.997} & 0.949 & 0.725 & 0.383 \\
            \hline
        \end{tabular} 
    \end{center}
\caption{ROC AUC for the distances based on SSIM, FSIM, Deep Learning and Procrustes}
\label{tab:resaucroc}
\end{table}

\begin{table}[b]
    \footnotesize
    \begin{center}
        \begin{tabular}{ccccc} 
            \hline 
            DS  & SSIM   & FSIM & Procrustes & VGG \\
            \hline
            1 & 1.0 & 1.0 & 1.0 & 0.006 \\
            2 & 1.0 & 1.0 & 1.0 & 0.017 \\
            3 & 0.901 &	\textbf{0.902} & 0.901 & 0.117 \\
            4 & 1.0 & 1.0 & 0.561 & 0.003 \\
            5 & 1.0 & 1.0 & 1.0 & 0.003 \\
            6 & 1.0 & 1.0 & 1.0 & 0.011 \\
            7 & 1.0 & 1.0 & 1.0 & 0.023 \\
            8 & \textbf{0.982} & 0.883 & 0.545 & 0.155 \\
            \hline
        \end{tabular} 
    \end{center}
\caption{PR AUC for the distances based on SSIM, FSIM, Procrustes and Deep Learning}
\label{tab:resaucpr}
\end{table}

The results show that more work is needed to use a pre-trained network to extract features and compute a distance that would encode the dissimilarity of printing on coins.
Further study would explore, among other things, the choice of network, the layer to be extracted, the pre-training dataset, or even the distance between feature vectors to be used, which could specifically extract this information. Tables \ref{tab:resaucroc} and \ref{tab:resaucpr} as well as ROC and PR curves in Fig.\ \ref{fig:rocr} show the similar performance of the SSIM- and FSIM-based distances. However, one major advantage of the first one is its lower computation time.

\begin{figure*}
\begin{center}
\includegraphics[scale=0.4]{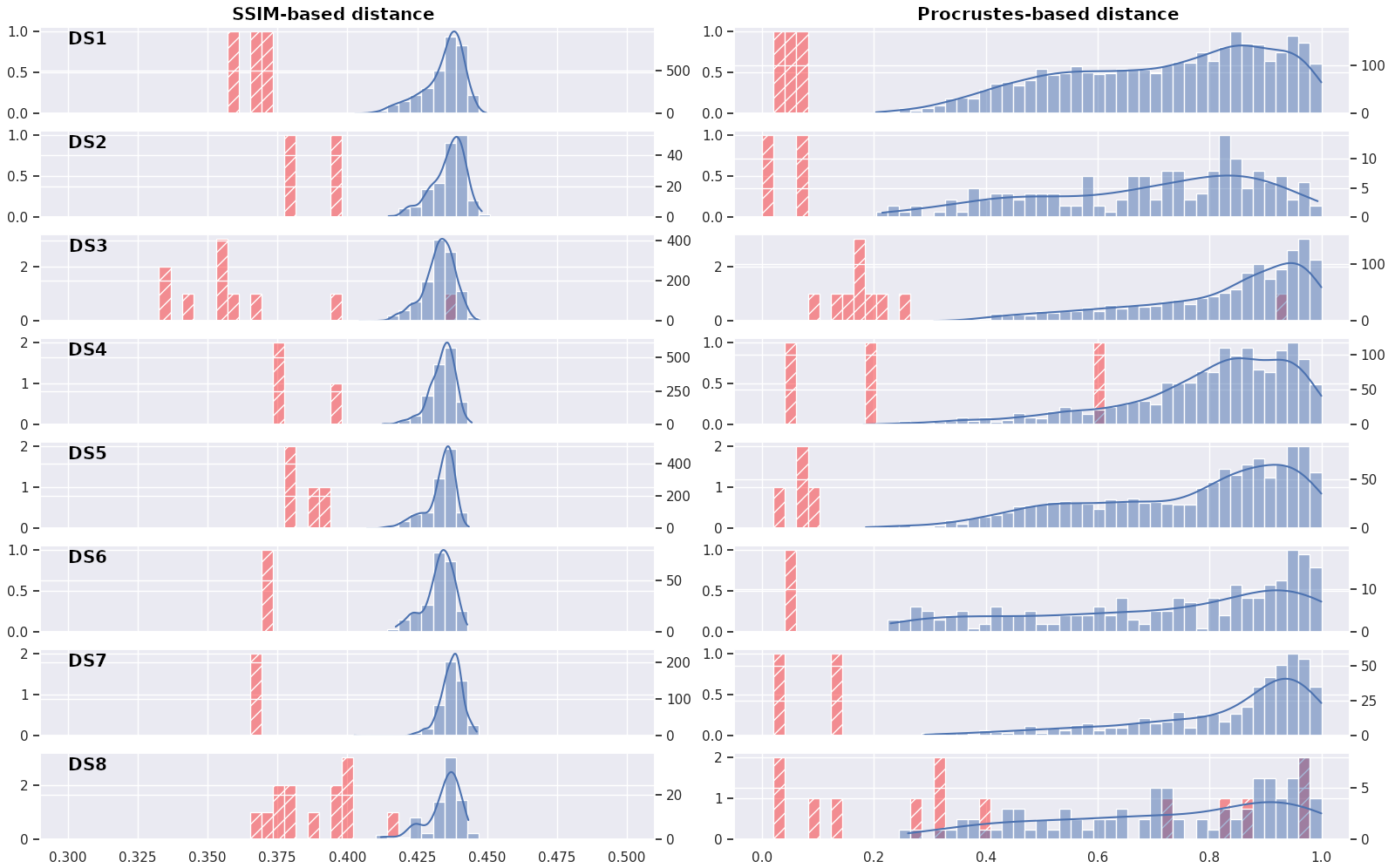}        
\end{center}
\caption{Histograms of the SSIM-based (left) and Procrustes-based (right) distance values. The bars in red with stripes correspond to the distances between images from the same cluster (y-axis scale is on the left), and the blue ones are inter-cluster distances (y-axis scale on the right).}
\label{fig:hist}
\end{figure*}

The remainder of this evaluation focuses on the SSIM-based distance and the Procrustes-based distance. The FSIM-based distance is no longer considered, since its results are similar to those of the SSIM-based distance, and the VGG-based distance is abandoned for lack of satisfactory results.
The histograms of the distances are shown in Fig.\ \ref{fig:hist}. We can see that the SSIM-based distance clearly separates intra-cluster distances from inter-cluster distances, except for Datasets 3 and 8 where only one intra-cluster distance is higher than the minimum inter-cluster distance.
With regard to the Procrustes-based distance, we note that one additional dataset observes this problem (Dataset 4), which now occurs 11 times, mainly in Dataset 8. The special feature of this dataset is that it contains more links than all the others (13 links), but is also the smallest of all (17 coins), see Table \ref{tab:dsslc}.

Table \ref{tab:res} provides the evaluation of three clustering predictions coming from two clustering techniques, namely \textit{Agglomerative Clustering with single linkage} (AC) and \textit{Bayesian Distance Clustering including both Cohesion and Repulsion terms in the likelihood} (CoRe) from \cite{Cohesion}. While the latter does not need any threshold to be defined beforehand, this is the case for AC. To this end, given a dataset, we use the other ones to estimate the best threshold, just like in the Leave-One-Out cross-validation procedure: once the optimal thresholds for each of the other datasets have been computed using the ground truth, several decision thresholds can be defined for the selected dataset: the maximum, the mean, the median, and the minimum of the optimal thresholds computed using the other datasets, resulting in the $AC_{max}$, $AC_{mean}$, $AC_{med}$ and $AC_{min}$ clustering techniques (see Table \ref{tab:res} for the first two, and supplementary material for the others).
Note that the use of other linkages with Agglomerative Clustering (\textit{complete} and \textit{average} linkages) lead to the same results, and the other clustering techniques tested (k-means, k-medoids, and CoRe without repulsion term) resulted in very poor clustering predictions. The presented results were computed using Scikit-learn \cite{scikit-learn} and the package provided with the paper on CoRe \cite{Cohesion}.

\begin{table*}
\footnotesize
\caption{Clustering performances (SSIM vs Procruste-based distance)}
\label{tab:res}
\centering
\begin{filecontents*}{results_latex_2.csv}
DS,clustering,Sssim-ARI,Sssim-NMI,Sssim-precision,Sssim-recall,Sssim-f1score,Sssim-accuracy,Sprocruste-ARI,Sprocruste-NMI,Sprocruste-precision,Sprocruste-recall,Sprocruste-f1score,Sprocruste-accuracy
1,$AC_{max}$,\textbf{1.0},\textbf{1.0},\textbf{1.0},1.0,\textbf{1.0},\textbf{1.0},0.008,0.771,0.005,1.0,0.009,0.805
2,,\textbf{1.0},\textbf{1.0},\textbf{1.0},1.0,\textbf{1.0},\textbf{1.0},0.251,0.903,0.154,1.0,0.267,0.936
3,,\textbf{0.947},\textbf{0.997},\textbf{1.0},0.9,\textbf{0.947},\textbf{0.999},0.856,0.992,0.818,0.9,0.857,0.998
4,,\textbf{1.0},\textbf{1.0},\textbf{1.0},\textbf{1.0},\textbf{1.0},\textbf{1.0},0.126,0.944,0.071,0.667,0.129,0.982
5,,\textbf{1.0},\textbf{1.0},\textbf{1.0},1.0,\textbf{1.0},\textbf{1.0},0.195,0.921,0.111,1.0,0.2,0.973
6,,\textbf{1.0},\textbf{1.0},\textbf{1.0},1.0,\textbf{1.0},\textbf{1.0},0.023,0.671,0.016,1.0,0.032,0.736
7,,\textbf{1.0},\textbf{1.0},\textbf{1.0},1.0,\textbf{1.0},\textbf{1.0},0.569,0.98,0.4,1.0,0.571,0.994
8,,\textbf{0.909},\textbf{0.973},\textbf{1.0},\textbf{0.846},\textbf{0.917},\textbf{0.985},0.332,0.856,0.5,0.308,0.381,0.904
1,$AC_{mea}$,1.0,1.0,1.0,1.0,1.0,1.0,1.0,1.0,1.0,1.0,1.0,1.0
2,,0.664,0.987,1.0,0.5,0.667,0.994,\textbf{1.0},\textbf{1.0},1.0,\textbf{1.0},\textbf{1.0},\textbf{1.0}
3,,\textbf{0.888},\textbf{0.993},1.0,\textbf{0.8},\textbf{0.889},\textbf{0.999},0.181,0.971,1.0,0.1,0.182,0.993
4,,\textbf{0.8},\textbf{0.997},1.0,\textbf{0.667},\textbf{0.8},0.999,0.5,0.994,1.0,0.333,0.5,0.999
5,,0.666,0.995,1.0,0.5,0.667,0.998,\textbf{1.0},\textbf{1.0},1.0,\textbf{1.0},\textbf{1.0},\textbf{1.0}
6,,1.0,1.0,1.0,1.0,1.0,1.0,1.0,1.0,1.0,1.0,1.0,1.0
7,,1.0,1.0,1.0,1.0,1.0,1.0,1.0,1.0,1.0,1.0,1.0,1.0
8,,\textbf{0.803},\textbf{0.948},1.0,\textbf{0.692},\textbf{0.818},\textbf{0.971},0.352,0.881,1.0,0.231,0.375,0.926
1,CoRe,\textbf{0.016},\textbf{0.781},\textbf{0.009},1.0,\textbf{0.017},\textbf{0.895},0.007,0.72,0.005,1.0,0.009,0.801
2,,-0.023,\textbf{0.523},0.0,0.0,0.0,\textbf{0.532},\textbf{0.0},0.0,\textbf{0.012},\textbf{1.0},\textbf{0.023},0.012
3,,\textbf{0.268},\textbf{0.933},\textbf{0.164},0.9,\textbf{0.277},\textbf{0.966},0.065,0.77,0.041,0.9,0.078,0.845
4,,\textbf{0.042},\textbf{0.857},\textbf{0.023},\textbf{1.0},\textbf{0.045},\textbf{0.918},0.025,0.796,0.015,0.667,0.029,0.913
5,,0.046,0.829,0.027,1.0,0.053,0.878,\textbf{0.072},\textbf{0.844},\textbf{0.041},1.0,\textbf{0.078},\textbf{0.92}
6,,\textbf{0.11},\textbf{0.912},\textbf{0.062},1.0,\textbf{0.118},\textbf{0.935},0.038,0.734,0.024,1.0,0.047,0.823
7,,\textbf{0.441},\textbf{0.98},\textbf{0.286},1.0,\textbf{0.444},\textbf{0.99},0.062,0.841,0.036,1.0,0.069,0.891
8,,\textbf{0.788},\textbf{0.949},\textbf{0.684},\textbf{1.0},\textbf{0.813},\textbf{0.956},0.123,0.743,0.182,0.308,0.229,0.801
\end{filecontents*}
\csvreader[
  tabular=|c|c|c|c|c|c|c|c||c|c|c|c|c|c|,
  table head= \multicolumn{2}{c}{} & \multicolumn{6}{c}{  \textbf{SSIM-based distance}} & \multicolumn{6}{c}{  \textbf{Procrustes-based distance}} \\
  \hline   Clust. &  ds &  ARI & NMI & Prec. & Rec. & $F_1$ & Acc. & ARI & NMI & Prec. & Rec. & $F_1$ & Acc.,
    late after line=,
    before line=\ifthenelse{\equal{\csvcolii}{} \OR \equal{\csvcolii}{mea} \OR \equal{\csvcolii}{med} \OR \equal{\csvcolii}{min} \OR \equal{\csvcolii}{max} }{\\}{\\\hline},
    table foot = \\\hline,
]{results_latex_2.csv}{}
{\csvcolii & \csvcoli & \csvcoliii & \csvcoliv & \csvcolv & \csvcolvi & \csvcolvii & \csvcolviii & \csvcolix & \csvcolx & \csvcolxi & \csvcolxii & \csvcolxiii & \csvcolxiv}
\end{table*}

Regarding prediction performance evaluation scores,
four binary classification scores are used, namely the Precision (Prec.), Recall (Rec.), $F_1$-score ($F_1$) and Accuracy (Acc.). Two additional clustering scores are also computed, namely the Ajusted Rand Index (ARI) and the Normalized Mutual Information (NMI). 
Looking at the results in Table \ref{tab:res}, we can see that the SSIM-based distance produces better results overall than the Procrustes-based method, and that the $AC_{max}$ clustering technique leads to results that take advantage of the full identification power of this distance. Note also that the SSIM-based distance with $AC_{max}$, $AC_{mea}$ and $AC_{med}$ produces a perfect precision on all datasets, \textit{i.e.} they don't produce any false positive.

The high clustering performance of $AC_{max}$ can be understood by looking at the histograms in Fig.\ \ref{fig:hist}: in this micro-clustering context \cite{betancourt2016flexible, betancourt2022random,Base, Cohesion}, some datasets contain very few die links, preventing an accurate estimation of the distribution of the distance values related to die links. However, it appears in this figure that the distribution of distance values that are not associated with die links have a stable lower bound across datasets. Since distributions are fairly well separated, the use of the maximum threshold aggregation strategy in AC allows to learn more efficiently than the other strategies the upper bound of the distribution associated with the die links. Indeed, the bias introduced by learning the threshold on highly unbalanced datasets can be lowered by using this strategy, taking advantage of datasets with distance values associated to die links closer to the lower bound of the other distribution (blue in Fig.\ \ref{fig:hist}). By using 
another aggregation strategy for the threshold definition with AC, the resulting threshold is not close enough to the lower bound of the distribution representing link absences (blue), and results in a lower recall.

Surprisingly, although more sophisticated, and better adapted to the problem (i.e. microclustering), CoRe \cite{Cohesion} doesn't offer the best clustering performance on these datasets. By adding some false positives, it also degrades the prediction for Dataset 7, that is perfectly clustered by any other method (with SSIM). However, of all the methods, CoRe offers the best recall on all datasets: this is a great quality for numismatists who prefer false positives to false negatives to help them in their investigations.


\section{Perspectives}
The performance scores presented in the previous section suggest that even more impressive results could be achieved by optimizing the parameters of the SSIM-based pipeline. Indeed, we recall here that the evaluation of our SSIM-based distance computation procedure was carried out with the default parameters of the functions proposed in the libraries used. More refined SSIM indices should also be studied in this context, such as Multi-scale SSIM (MS-SSIM, \cite{wang2003multiscale}), Complex Wavelet SSIM (CW-SSIM, \cite{sampat2009complex}) or DISTS \cite{ding2020image}.
In this paper, the structured similarity (SSIM) and the feature similarity (FSIM) indices have demonstrated an equivalent detection performance, although the SSIM-based distance was faster to compute.

While this aspect has not been analysed precisely, the computation time is a great advantage of the SSIM-based distance: it takes a few hours to compute it on all the datasets, while the computation of the Procrustes-based one takes more than a day. This significant result means that the system is now ready for online production, enabling the analysis of new databases from all over the world. The performance of this approach also makes it possible to consider the creation of human-machine interfaces, for instance displaying the full output of SSIM (on the right in Fig.\ \ref{fig:ssim_steps_and_result_exemple}), to enhance the processing capabilities of numismatists, or providing the list of coin pairs in ascending order of SSIM-based distance, to let them focus on the most likely links first.
These techniques will be used on the other (numerous) datasets from the presented treasure. 

\section{Conclusion}
This paper presents the first dataset of images of ancient coins made available online, and labeled for the challenge of coin die link detection \cite{AFRCBK_2024}. This dataset provides the scientific community with the opportunity to benchmark Computer Vision solutions to this problem. 
Moreover, a new procedure for computing a distance between coin pictures, based on SSIM, is proposed. This pipeline, as well as other pipelines of the literature 
are evaluated by various means: histograms, ROC and PR curves, as well as results from distance-based clustering algorithms. Decision threshold learning on validation datasets yields near-perfect results when using a maximum-based threshold aggregation strategy. This impressive performance makes possible the automatic analysis of image databases of ancient coins for die link detection, which will allow in the future the extraction of crucial historical information in a more systematic way.

{\small
\balance
\bibliographystyle{plainurl}
\bibliography{accadil}
}

\end{document}


\title{Supplementary Material\\
\small A High-Accuracy SSIM-based Scoring System for Coin Die Link Identification}

\author{
}
\maketitle

\section{Datasets}
\begin{figure}
    \centering
	  \includegraphics[width=0.7\linewidth]{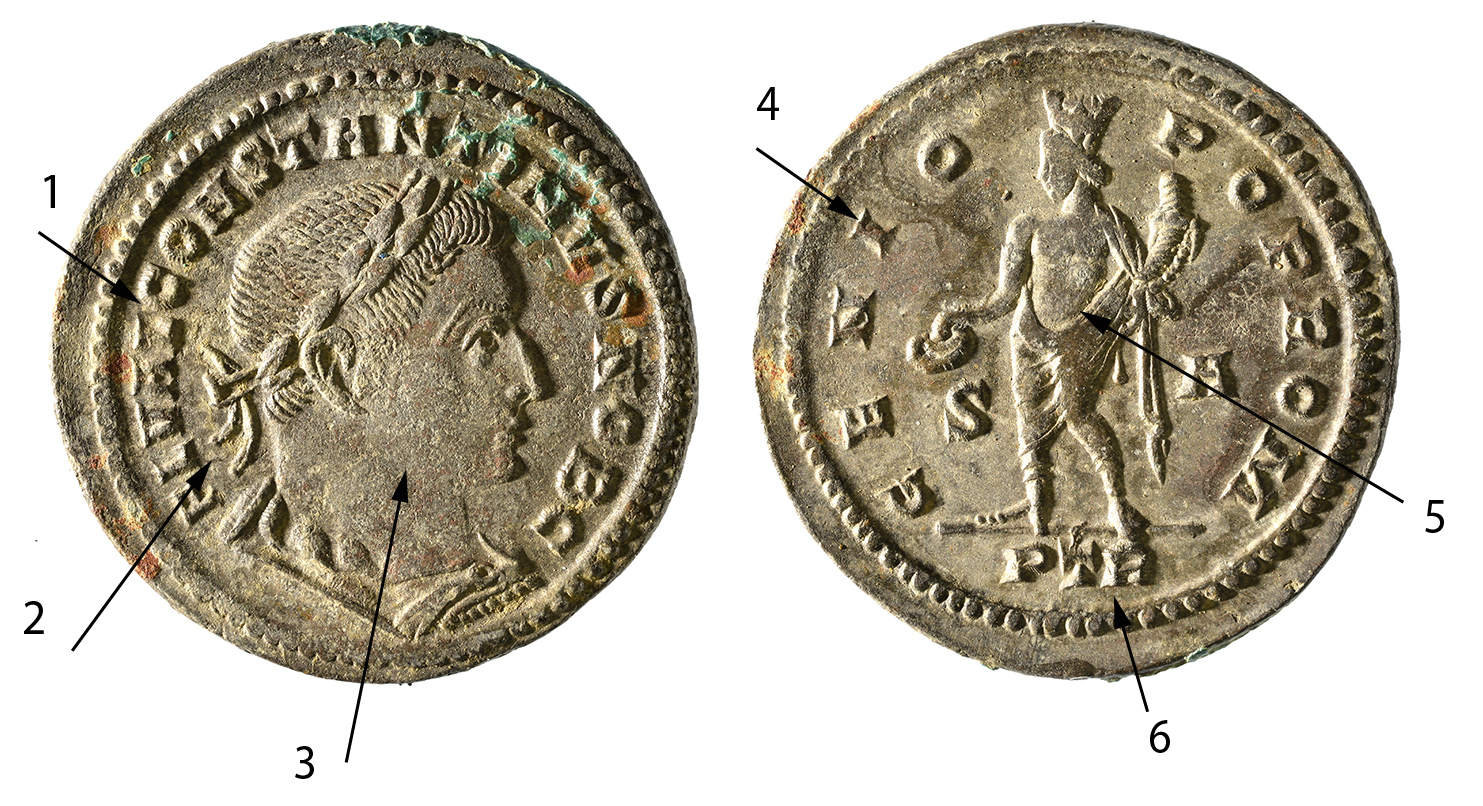}
	  \caption{1) Obverse legend = \textbf{FL VAL CONSTANTINVS NOB C}; 2) Ribbon code = \textbf{3} (i.e. 2 vertical ribbons); 3) Bust code = \textbf{A*2} (i.e. bust laureate, draped, cuirassed, right, view from rear); 4) Obverse legend = \textbf{GENIO POP ROM}; 5) Reverse code = \textbf{genio 6} (i.e. \textit{Genius}, turreted, draped, standing left, holding \textit{patera} in right hand and \textit{cornucopiae} in left hand); 6) Mint mark = \textbf{S $|$ A // PTR} (i.e “S$|$A” emission, struck at \textit{Prima officina}, Treveri mint)}
	  \label{fig:coin_6_carac} 
\end{figure}

The Juillac treasure was discovered in 2011 in the municipality of L'Isle-Jourdain (Gers, France). The datasets used for our work come from the scientific study of this important treasure. It contains more than 23,200 Roman coins, mainly dated between 294 and 313 AD. The archaeologists and numismatists studying this hoard analyzed each coin, which is documented on both sides (called the obverse and reverse) with a digital photograph and several descriptive headings, six of which are used for this research. For the obverse, there are three headings: the text of the legend engraved around the portrait of the emperor (his name and titles), the bust code (he can be draped, armoured, bareheaded, with headdress, on the left, on the right, etc.), and the ribbon code which specifies the type of attachment for the crown worn by the emperor. On the reverse, these are: the text of the legend engraved around the figure represented (often the name of a divinity or allegory), the reverse code (describing the divinity or allegory), and the mark of the mint that produced the coin (London, Trier, Lyon, Rome, Carthage, etc.). An example is given in the 
Fig. \ref{fig:coin_6_carac}). 

These six headings alone make it possible to classify all the coins by type of obverse and type of reverse. If we only keep the types composed of at least two coins, the database thus contains 658 different types of reverse (from 2 to 1 395 coins), and 379 different types of obverse (from 2 to 1 255 coins). For the study of this hoard, the numismatists created an innovative database in the field of large hoards. It allows easy access to the record of each coin and, more importantly, to the coins of each previously identified type, enabling comparisons between pairs of coins. The visual analysis of die links has thus started for certain types of obverse or reverse. At the time of our work, this is the case for coins from the \textit{Ticinum} mint, with a relatively small number of coins examined (lots numbering from 2 to 93). Eight sets are used as references, called here DS1, DS2, ..., DS8. Their types are listed in Table \ref{tab:tab_datasets_informations_numism}. The numismatists allowed us to use and publicly share these datasets.

\begin{table}
\caption{Numismatic information about the datasets}
\footnotesize
    \begin{center}
        \begin{tabular}{ccc} 
            \hline 
            Dataset  & Type Legend   & Mint mark      \\
            \hline
            DS1             &  GENIO POPV-LI ROMANI               &  * | - // ST  \\
            DS2             &  GENIO POPV-LI ROMANI               &  - | - // T \\
            DS3             &  PROVIDENTIA DEORVM QVIES AVGG      &  - | • // TT \\
            DS4             &  SACRA MONET AVGG - ET CAESS NOSTR  &  - | - // ST• \\
            DS5             &  SACRA MONET AVGG ET CAESS NOSTR    &  - | * // TT \\
            DS6             &  SACRA MONET AVGG ET CAESS NOSTR    &  - | V // AQP \\
            DS7             &  VIRTVS AV-GG ET CAESS NN           &  - | - // AQ$\Gamma$ \\
            DS8             &  VIRTVS AV-GG ET CAESS NN           &  A | - // PT \\
            \hline
        \end{tabular}
        \label{tab:tab_datasets_informations_numism}
    \end{center}
\end{table}

\section{Clustering Performances}
Table \ref{tab:res} shows all the results obtained for the 5 best-performing clustering predictions coming from two clustering techniques, namely \textit{Agglomerative Clustering with single linkage} (AC) and \textit{Bayesian Distance Clustering including both Cohesion and Repulsion terms in the likelihood} (CoRe) from \cite{Cohesion}.

\begin{table*}[t]
\footnotesize
\caption{Clustering performances (SSIM vs Procruste-based distance)}
\label{tab:res}
\centering
\begin{filecontents*}{results_latex.csv}
DS,clustering,Sssim-ARI,Sssim-NMI,Sssim-precision,Sssim-recall,Sssim-f1score,Sssim-accuracy,Sprocruste-ARI,Sprocruste-NMI,Sprocruste-precision,Sprocruste-recall,Sprocruste-f1score,Sprocruste-accuracy
1,$AC_{max}$,\textbf{1.0},\textbf{1.0},\textbf{1.0},1.0,\textbf{1.0},\textbf{1.0},0.008,0.771,0.005,1.0,0.009,0.805
2,,\textbf{1.0},\textbf{1.0},\textbf{1.0},1.0,\textbf{1.0},\textbf{1.0},0.251,0.903,0.154,1.0,0.267,0.936
3,,\textbf{0.947},\textbf{0.997},\textbf{1.0},0.9,\textbf{0.947},\textbf{0.999},0.856,0.992,0.818,0.9,0.857,0.998
4,,\textbf{1.0},\textbf{1.0},\textbf{1.0},\textbf{1.0},\textbf{1.0},\textbf{1.0},0.126,0.944,0.071,0.667,0.129,0.982
5,,\textbf{1.0},\textbf{1.0},\textbf{1.0},1.0,\textbf{1.0},\textbf{1.0},0.195,0.921,0.111,1.0,0.2,0.973
6,,\textbf{1.0},\textbf{1.0},\textbf{1.0},1.0,\textbf{1.0},\textbf{1.0},0.023,0.671,0.016,1.0,0.032,0.736
7,,\textbf{1.0},\textbf{1.0},\textbf{1.0},1.0,\textbf{1.0},\textbf{1.0},0.569,0.98,0.4,1.0,0.571,0.994
8,,\textbf{0.909},\textbf{0.973},\textbf{1.0},\textbf{0.846},\textbf{0.917},\textbf{0.985},0.332,0.856,0.5,0.308,0.381,0.904
1,$AC_{mea}$,1.0,1.0,1.0,1.0,1.0,1.0,1.0,1.0,1.0,1.0,1.0,1.0
2,,0.664,0.987,1.0,0.5,0.667,0.994,\textbf{1.0},\textbf{1.0},1.0,\textbf{1.0},\textbf{1.0},\textbf{1.0}
3,,\textbf{0.888},\textbf{0.993},1.0,\textbf{0.8},\textbf{0.889},\textbf{0.999},0.181,0.971,1.0,0.1,0.182,0.993
4,,\textbf{0.8},\textbf{0.997},1.0,\textbf{0.667},\textbf{0.8},0.999,0.5,0.994,1.0,0.333,0.5,0.999
5,,0.666,0.995,1.0,0.5,0.667,0.998,\textbf{1.0},\textbf{1.0},1.0,\textbf{1.0},\textbf{1.0},\textbf{1.0}
6,,1.0,1.0,1.0,1.0,1.0,1.0,1.0,1.0,1.0,1.0,1.0,1.0
7,,1.0,1.0,1.0,1.0,1.0,1.0,1.0,1.0,1.0,1.0,1.0,1.0
8,,\textbf{0.803},\textbf{0.948},1.0,\textbf{0.692},\textbf{0.818},\textbf{0.971},0.352,0.881,1.0,0.231,0.375,0.926
1,$AC_{med}$,1.0,1.0,1.0,1.0,1.0,1.0,1.0,1.0,1.0,1.0,1.0,1.0
2,,0.664,0.987,1.0,0.5,0.667,0.994,\textbf{1.0},\textbf{1.0},1.0,\textbf{1.0},\textbf{1.0},\textbf{1.0}
3,,\textbf{0.888},\textbf{0.993},1.0,\textbf{0.8},\textbf{0.889},\textbf{0.999},0.181,0.971,1.0,0.1,0.182,0.993
4,,\textbf{0.8},\textbf{0.997},1.0,\textbf{0.667},\textbf{0.8},0.999,0.5,0.994,1.0,0.333,0.5,0.999
5,,1.0,1.0,1.0,1.0,1.0,1.0,1.0,1.0,1.0,1.0,1.0,1.0
6,,1.0,1.0,1.0,1.0,1.0,1.0,1.0,1.0,1.0,1.0,1.0,1.0
7,,\textbf{1.0},\textbf{1.0},1.0,\textbf{1.0},\textbf{1.0},\textbf{1.0},0.666,0.994,1.0,0.5,0.667,0.998
8,,\textbf{0.803},\textbf{0.948},1.0,\textbf{0.692},\textbf{0.818},\textbf{0.971},0.352,0.881,1.0,0.231,0.375,0.926
1,$AC_{min}$,\textbf{0.8},\textbf{0.998},1.0,\textbf{0.667},\textbf{0.8},\textbf{1.0},0.5,0.996,1.0,0.333,0.5,0.999
2,,0.0,0.975,0.0,0.0,0.0,0.988,\textbf{0.664},\textbf{0.987},\textbf{1.0},\textbf{0.5},\textbf{0.667},\textbf{0.994}
3,,\textbf{0.888},\textbf{0.993},\textbf{1.0},\textbf{0.8},\textbf{0.889},\textbf{0.999},0.0,0.968,0.0,0.0,0.0,0.993
4,,0.0,0.991,0.0,0.0,0.0,0.998,\textbf{0.5},\textbf{0.994},\textbf{1.0},\textbf{0.333},\textbf{0.5},\textbf{0.999}
5,,0.0,0.988,0.0,0.0,0.0,0.997,\textbf{0.399},\textbf{0.991},\textbf{1.0},\textbf{0.25},\textbf{0.4},0.997
6,,0.0,0.99,0.0,0.0,0.0,0.996,\textbf{1.0},\textbf{1.0},\textbf{1.0},\textbf{1.0},\textbf{1.0},\textbf{1.0}
7,,\textbf{1.0},\textbf{1.0},1.0,\textbf{1.0},\textbf{1.0},\textbf{1.0},0.666,0.994,1.0,0.5,0.667,0.998
8,,0.131,0.851,1.0,0.077,0.143,0.912,\textbf{0.247},\textbf{0.866},1.0,\textbf{0.154},\textbf{0.267},\textbf{0.919}
1,CoRe,\textbf{0.016},\textbf{0.781},\textbf{0.009},1.0,\textbf{0.017},\textbf{0.895},0.007,0.72,0.005,1.0,0.009,0.801
2,,-0.023,\textbf{0.523},0.0,0.0,0.0,\textbf{0.532},\textbf{0.0},0.0,\textbf{0.012},\textbf{1.0},\textbf{0.023},0.012
3,,\textbf{0.268},\textbf{0.933},\textbf{0.164},0.9,\textbf{0.277},\textbf{0.966},0.065,0.77,0.041,0.9,0.078,0.845
4,,\textbf{0.042},\textbf{0.857},\textbf{0.023},\textbf{1.0},\textbf{0.045},\textbf{0.918},0.025,0.796,0.015,0.667,0.029,0.913
5,,0.046,0.829,0.027,1.0,0.053,0.878,\textbf{0.072},\textbf{0.844},\textbf{0.041},1.0,\textbf{0.078},\textbf{0.92}
6,,\textbf{0.11},\textbf{0.912},\textbf{0.062},1.0,\textbf{0.118},\textbf{0.935},0.038,0.734,0.024,1.0,0.047,0.823
7,,\textbf{0.441},\textbf{0.98},\textbf{0.286},1.0,\textbf{0.444},\textbf{0.99},0.062,0.841,0.036,1.0,0.069,0.891
8,,\textbf{0.788},\textbf{0.949},\textbf{0.684},\textbf{1.0},\textbf{0.813},\textbf{0.956},0.123,0.743,0.182,0.308,0.229,0.801
\end{filecontents*}
\csvreader[
  tabular=|c|c|c|c|c|c|c|c||c|c|c|c|c|c|,
  table head= \multicolumn{2}{c}{} & \multicolumn{6}{c}{  \textbf{SSIM-based distance}} & \multicolumn{6}{c}{  \textbf{Procrustes-based distance}} \\
  \hline   Clust. &  ds &  ARI & NMI & Prec. & Rec. & $F_1$ & Acc. & ARI & NMI & Prec. & Rec. & $F_1$ & Acc.,
    late after line=,
    before line=\ifthenelse{\equal{\csvcolii}{} \OR \equal{\csvcolii}{mea} \OR \equal{\csvcolii}{med} \OR \equal{\csvcolii}{min} \OR \equal{\csvcolii}{max} }{\\}{\\\hline},
    table foot = \\\hline,
]{results_latex.csv}{}
{\csvcolii & \csvcoli & \csvcoliii & \csvcoliv & \csvcolv & \csvcolvi & \csvcolvii & \csvcolviii & \csvcolix & \csvcolx & \csvcolxi & \csvcolxii & \csvcolxiii & \csvcolxiv}
\end{table*}

While the latter does not need any threshold to be defined beforehand, this is the case for AC. To this end, given a dataset, we use the other ones to estimate the best threshold, just like in the Leave-One-Out cross-validation procedure: once the optimal thresholds for each of the other datasets have been computed using the ground truth, several decision thresholds can be defined for the selected dataset: the maximum, the mean, the median, and the minimum of the optimal thresholds computed using the other datasets, resulting in the $AC_{max}$, $AC_{mean}$, $AC_{med}$ and $AC_{min}$ clustering techniques (see Table \ref{tab:res}).
Note that the use of other linkages with Agglomerative Clustering (\textit{complete} and \textit{average} linkages) lead to the same results, and the other clustering techniques tested (k-means, k-medoids, and CoRe without repulsion term) resulted in very poor clustering predictions. The presented results were computed using Scikit-learn \cite{scikit-learn} and the package provided with the paper on CoRe \cite{Cohesion}.

Agglomerative Clustering with single linkage gets the best results with the max threshold aggregation strategy ($AC_{max}$), then with the mean ($AC_{mea}$) and median ($AC_{med}$) threshold aggregation strategies (which have similar results), and finally with the min threshold aggregation strategy ($AC_{min}$), which performs as well as $CoRe$.

{\small
\bibliographystyle{plainurl}
\bibliography{accadil}